\documentclass[10pt, onecolumn, a4paper]{article}

\usepackage[verbose=true,letterpaper]{geometry}
\makeatletter
\AtBeginDocument{
  \newgeometry{
    textheight=9in,
    textwidth=5.5in,
    top=1in,
    headheight=12pt,
    headsep=25pt,
    footskip=30pt
  }
}

\makeatother
\renewenvironment{abstract}%
{%
  \vskip 0.075in%
  \centerline%
  {\large\bf Abstract}%
  \vspace{0.5ex}%
  \begin{quote}%
}
{
  \par%
  \end{quote}%
  \vskip 1ex%
}

\usepackage{palatino}
\usepackage[utf8]{inputenc} 
\usepackage[T1]{fontenc}    
\usepackage{xcolor}
\usepackage{hyperref}
\ifdefined\nohyperref\else\ifdefined\hypersetup
  \definecolor{mydarkblue}{rgb}{0,0.08,0.45}
  \hypersetup{ %
    pdftitle={},
    pdfauthor={},
    pdfsubject={},
    pdfkeywords={},
    pdfborder=0 0 0,
    pdfpagemode=UseNone,
    colorlinks=true,
    linkcolor=mydarkblue,
    citecolor=mydarkblue,
    filecolor=mydarkblue,
    urlcolor=mydarkblue,
    pdfview=FitH}
\usepackage{url}            
\usepackage{booktabs}       
\usepackage{amsfonts}       
\usepackage{nicefrac}       
\usepackage{microtype}      
\usepackage{calc}
\usepackage{bm}
\usepackage{amssymb,amsmath}
\usepackage{mathtools}
\mathtoolsset{showonlyrefs}
\usepackage{wrapfig}
\usepackage{color}
\usepackage{caption}
\usepackage{soul}

\usepackage[compress]{natbib}

\usepackage{graphicx}
\usepackage{subfigure}
\usepackage{amsthm}

\usepackage{algorithm}
\usepackage{algpseudocode}
\usepackage{amsmath}
\usepackage{amssymb}
\usepackage{xcolor}
\usepackage{mathtools}
\usepackage{sidecap}
\usepackage{wrapfig}

\usepackage{enumitem}
\setlist[itemize,enumerate]{leftmargin=*}

\usepackage{mdframed}
\definecolor{theoremcolor}{rgb}{0.94, 0.94, 0.94}
\definecolor{examplecolor}{rgb}{1, 1, 1.0}
\mdfsetup{
    backgroundcolor=theoremcolor,
    linewidth=0pt,
}

\usepackage{xfrac}
\usepackage{comment}

\newmdtheoremenv[linewidth=0.5pt,innerleftmargin=4pt,innerrightmargin=4pt]{definition}{Definition}
\newmdtheoremenv[linewidth=0.5pt,innerleftmargin=4pt,innerrightmargin=4pt]{proposition}{Proposition}
\newmdtheoremenv[linewidth=0.5pt,innerleftmargin=4pt,innerrightmargin=4pt]{example}{Example}
\newmdtheoremenv[linewidth=0.5pt,innerleftmargin=4pt,innerrightmargin=4pt]{corollary}{Corollary}
\newmdtheoremenv[linewidth=0.5pt,innerleftmargin=4pt,innerrightmargin=4pt]{theorem}{Theorem}
\newmdtheoremenv[linewidth=0.5pt,innerleftmargin=4pt,innerrightmargin=4pt]{lemma}{Lemma}
\newmdtheoremenv[linewidth=0.5pt,innerleftmargin=4pt,innerrightmargin=4pt]{remark}{Remark}

\DeclareMathOperator*{\argmax}{arg\,max}

\newcommand{\xx}[1]{{x^{(#1)}}}
\newcommand{\yy}[1]{{y^{(#1)}}}
\newcommand{\yyhat}[1]{{\hat{y}^{(#1)}}}

\newcommand{\xxtilde}[1]{{{\tilde{x}}^{(#1)}}}
\newcommand{\xxtildet}[1]{{{\tilde{x}}^{(#1)^\top}}}

\usepackage{fancyhdr}
\usepackage{titling}
\usepackage{authblk}

\usepackage{inconsolata}
\usepackage{tikz}

\usepackage{xspace}

\title{\bf Learning with the $p$-adics\\[2ex]}

\author{Andr\'e F.~T.~Martins\textsuperscript{1,2}}

\affil{\textsuperscript{1}%
  \normalsize{Instituto Superior Técnico, Universidade de Lisboa, Portugal}
  } 
\affil{
\textsuperscript{2}%
  \normalsize{Instituto de Telecomunica\c{c}\~oes, Lisboa, Portugal} 
  }

\affil{\normalsize{\url{andre.t.martins@tecnico.ulisboa.pt}}}

\date{}

\begin{document}

\maketitle

\begin{abstract}
Existing machine learning frameworks operate over the field of real numbers ($\mathbb{R}$) and learn representations in real (Euclidean or Hilbert) vector spaces (\textit{e.g.}, $\mathbb{R}^d$). 
Their underlying geometric properties align well with intuitive concepts such as linear separability, minimum enclosing balls, and subspace projection; and basic calculus provides a toolbox for learning through gradient-based optimization. 

But is this the only possible choice? 
In this paper, we study the suitability of a radically different field as an alternative to $\mathbb{R}$---the ultrametric and non-archimedean space of $p$-adic numbers,  $\mathbb{Q}_p$. 
The hierarchical structure of the $p$-adics and their interpretation as infinite strings make them an appealing tool for code theory and hierarchical representation learning. 
Our exploratory theoretical work establishes the building blocks for classification, regression, and representation learning with the $p$-adics, providing learning models and algorithms. 
We illustrate how simple Quillian semantic networks can be represented as a compact $p$-adic linear network, a construction which is not possible with the field of reals. 
We finish by discussing open problems and opportunities for future research enabled by this new framework.
\end{abstract}

\begin{figure}[h]
    \centering
    \includegraphics[width=0.6\columnwidth]{figures/Qp_hierarchy.pdf}%
    \caption{Hierarchical structure of $\mathbb{Q}_p$ for $p=2$.}
    \label{fig:Qp_hierarchy_cover}
\end{figure}

\clearpage
\tableofcontents
\clearpage

\section{Introduction}

Since they have been introduced by \citet{hensel1897neue}, $p$-adic numbers have seen numerous applications in number theory, algebraic geometry, physics, and other fields \citep{koblitz1984padic,robert2000course,gouvea2020padic}. 
They differ from the real numbers in many important ways, which leads to many fascinating and surprising results, such that the equality
$$1 + 2 + 4 + 8 + \ldots = -1$$
or the fun fact that in the $p$-adic world all triangles are isosceles and any point in a ball is a center of that ball. 

Yet, with only a few exceptions, very little work has investigated the potential of  $p$-adic numbers in machine learning. 
\citet{bradley2009p} studies clustering of $p$-adic data and proposes suboptimal algorithms for minimizing cluster energies. \citet{murtagh2004ultrametricity, murtagh2009data} analyze dendrograms and ultrametricity in data. 
\citet{chierchia2019ultrametric} and \citet{cohen2020efficient} develop procedures to fit ultrametrics to data. 
\citet{khrennikov1999learning} propose a ``$p$-adic neural network'' (similar to our unidimensional linear classifier in \S\ref{sec:classification}). 
\citet{baker2022number} use a variant of $p$-adic regression for a small sequence-to-sequence problem in natural language processing, bearing some similarity with our formulation in \S\ref{sec:regression}. 

This paper is an attempt to establish the foundations for $p$-adic machine learning by developing building blocks for classification (\S\ref{sec:classification}) and regression problems (\S\ref{sec:regression}) and exploring the expressive power of $p$-adic representations (\S\ref{sec:representation_learning}). 
The paper ends with a selection of open problems (\S\ref{sec:open_problems}) which I believe need to be addressed to make this framework practically useful. 
(The title is intentionally misleading: it is mostly about what \textit{I have been learning} with the $p$-adics, and not so much about \textit{how one can do machine learning} with the $p$-adics.) 

This endeavour comes with several challenges, since  
the classical tools of gradient-based optimization and statistics are not readily available in the world of the $p$-adics: although we can do calculus and find roots of functions using Newton's method, the topological properties of the $p$-adics seem to make derivatives not so useful (\textit{e.g.}, a function with zero derivative everywhere might not be constant), and it is not obvious how to optimize since the $p$-adics, unlike the reals, are not an ordered field. 
However, they possess a very interesting hierarchical structure which appears promising for representation learning and certain classification and regression problems. 

\paragraph{Notation.} We denote by $\mathbb{R}$ and $\mathbb{Q}$ the fields of real and rational numbers, respectively, $\mathbb{R}_+$ the non-negative reals, and $\mathbb{Z}$ the ring of integer numbers. We denote $[n] = \{1, ..., n\}$.

\section{Background}

We start by reviewing ultrametric spaces, the field of $p$-adic numbers, and the ring of $p$-adic integers, along with their basic properties \citep{gouvea2020padic}. 

\subsection{Non-Archimedean absolute values and ultrametrics}\label{sec:nonarch}

Let $\mathbb{K}$ be a field (such as $\mathbb{Q}$ or $\mathbb{R}$) and let $\mathbb{R}_+$ denote the non-negative real numbers. 
An \textbf{absolute value} on $\mathbb{K}$ is a function $|.|: \mathbb{K} \rightarrow \mathbb{R}_+$ satisfying  
(i) $|x|=0$ iff $x=0$; (ii) $|xy| = |x||y|$ for all $x,y \in \mathbb{K}$; (iii) $|x+y| \le |x|+|y|$ for all $x,y \in \mathbb{K}$. 
An absolute value is called \textbf{non-Archimedean} if it has the following property (stronger than iii): 
\begin{align}\label{eq:nonarchimedean}
    |x+y| &\le \max\{|x|, |y|\}, \quad \text{for all $x,y \in \mathbb{K}$}. 
\end{align}
It is called ``Archimedean'' otherwise.%
\footnote{The name comes from the fact that Archimedean absolute values satisfy the Archimedean property: for any $x,y \in \mathbb{K}$ with $x \ne 0$, there is an integer $k \in \mathbb{Z}$ such that $|kx| > |y|$, which is equivalent to the assertion that there are arbitrarily ``big'' integers, an observation which goes back to Archimedes. This does not happen with non-Archimedean absolute values, where \eqref{eq:nonarchimedean} implies $|kx| = |x + \ldots + x| \le |x|$.} %

\begin{example}[trivial absolute value]
The function defined as $|x| = 1$ if $x \ne 0$ and $|0|=0$ is an absolute value, called the trivial absolute value. It is non-Archimedean. 
\end{example}

\begin{example}[usual absolute value on $\mathbb{Q}$ and $\mathbb{R}$]
\label{ex:usual_abs}
Let $\mathbb{K} = \mathbb{Q}$ or $\mathbb{R}$. 
The usual absolute value $|x| := \max\{x, -x\}$ is Archimedean. 
\end{example}

\begin{example}[$p$-adic absolute value on $\mathbb{Q}$]\label{ex:padic_abs} 
Let $p$ be a prime number. 
Any nonzero $x \in \mathbb{Q}$ can be written uniquely as $x=p^{n}\frac{a}{b}$, where $a$ and $b$ are co-prime integers not divisible by $p$, and $n\in \mathbb{Z}$. 
The $p$-adic absolute value on $\mathbb{Q}$ is
\begin{align}
|x|_p := \left\{
\begin{array}{ll}
p^{-n} & \text{if $x \ne 0$}\\
0 & \text{if $x=0$}. 
\end{array}
\right.
\end{align}
This absolute value (used extensively throughout this paper) is non-Archimedean. 
\end{example}

An absolute value induces a \textbf{metric} (and consequently a topology) on $\mathbb{K}$ through the distance function $d(x, y) := |x-y|$. 
It results from properties (i--iii) above that we must have (i) $d(x,y) \ge 0$ $\forall x,y \in \mathbb{K}$, with equality iff $x=y$; (ii) $d(x,y) = d(y,x)$ $\forall x,y \in \mathbb{K}$; (iii) $d(x,z) \le d(x,y) + d(y,z)$ $\forall 
 x,y,z \in \mathbb{K}$. 
Property (iii) is called the (weak) triangle inequality. 
If $|\cdot|$ is non-Archimedean, then the induced metric satisfies the \textbf{strong triangle inequality}:
\begin{align}\label{eq:sti}
    d(x,z) \le \max\{d(x,y), d(y,z)\} \quad \text{$\forall x,y,z \in \mathbb{K}$}. 
\end{align}
Metrics satisfying \eqref{eq:sti} are called \textbf{ultrametrics} and the corresponding $\mathbb{K}$ is called a {ultrametric space}. Ultrametric spaces have the interesting property that \textbf{every triangle is isosceles}---more specifically, for any $x, y, z \in \mathbb{K}$, 
$d(x,y) \ne d(y,z)$ implies
$d(x,z) = \max\{d(x,y), d(y,z)\}$. We provide a simple proof in Appendix~\ref{sec:proof_isosceles}. %

\subsection{The field $\mathbb{Q}_p$ of $p$-adic numbers and the ring $\mathbb{Z}_p$ of $p$-adic integers}

Let us recall how 
the set of real numbers $\mathbb{R}$ is constructed 
by {completing} the set of rationals $\mathbb{Q}$. 
Given an absolute value and its induced metric, we can define open balls and a notion of convergence. 
A field $\mathbb{K}$ is \textbf{complete} if any Cauchy sequence%
\footnote{A sequence of elements $x_n \in \mathbb{K}$ is a Cauchy sequence if  $\forall \epsilon>0$ there is $M \in \mathbb{N}$ such that $m,n \ge M$ imply $|x_n-x_m| < \epsilon$.} %
in $\mathbb{K}$ converges to a limit point in $\mathbb{K}$---for example, $\mathbb{Q}$ is \textit{not} complete with respect to the usual absolute value.%
\footnote{For example, the sequence defined by $x_{n+1} = \tfrac{1}{2}x_n + x_n^{-1}$ 
with $x_n=1$ converges to $\sqrt{2} \notin \mathbb{Q}$  (this is the sequence obtained by applying Newton's method to find a solution of $x^2-2 = 0$).} %
By filling the ``holes'' in  $\mathbb{Q}$ with these limit points---the irrational numbers---we obtain the larger, complete field of real numbers $\mathbb{R}$. 

What happens if we follow this idea but use instead the $p$-adic absolute value $|\cdot|_p$ to complete $\mathbb{Q}$?%
\footnote{In fact, Ostrowski's theorem \citep[Theorem 3.1.4]{gouvea2020padic} remarkably states that the three examples in \S\ref{sec:nonarch} tell the full picture regarding absolute values in $\mathbb{Q}$:
every non-trivial absolute value on $\mathbb{Q}$ is equivalent (in the sense of inducing the same topology) 
to one of the $p$-adic absolute values or to the usual absolute value. 
This result suggests we should give the $p$-adic absolute values the same  importance we give to the usual absolute value. Denoting the latter by $|.|_\infty$, this is nicely captured in the product formula which relates \textit{all} the non-trivial absolute values:
$\prod_{p \le \infty} |x|_p = 1$ \citep[Proposition 3.1.5]{gouvea2020padic}. We come back to this beautiful formula in \S\ref{sec:adelic_predictors}.} %
$\mathbb{Q}$ is also incomplete with respect to the metric $d(x,y) := |x-y|_p$. 
By adding the limits of Cauchy sequences in $\mathbb{Q}$ with respect to this metric, we obtain 
the field of \textbf{$p$-adic numbers}, denoted by $\mathbb{Q}_p$. 
The subset $\mathbb{Z}_p := \{x \in \mathbb{Q}_p\,:\, |x|_p \le 1\} \subset \mathbb{Q}_p$ is called the set of \textbf{$p$-adic integers} and it has a ring structure. 

\begin{proposition}[$p$-adic expansion]
Every 
$x \in \mathbb{Z}_p$ can be written uniquely as an infinite ``digit'' expansion in base $p$:
\begin{align}
x &= a_0 + a_1p + ... + a_n p^n + ... = \textstyle \sum_{i=0}^\infty a_i p^i, 
\end{align}
where each $a_i \in \{0, ..., p-1\}$. 
When $p$ is clear from the context, we abbreviate this as 
$x = \textcolor{red}{\cdots a_2 a_1 a_0}.$ 

Every $x \in \mathbb{Q}_p$ can be written uniquely as:
\begin{align}\label{eq:padic_expansion}
x 
&= \textstyle \sum_{i=-m}^\infty a_i p^i := \textcolor{red}{\cdots a_2 a_1 a_0\,\,.\,\, a_{-1} \ldots a_{-m}},
\end{align}
where $m \in \mathbb{Z}$, each $a_i \in \{0, ..., p-1\}$, and $a_{-m} \ne 0$. 
Furthermore, we have $|x|_p = p^m$. 
\end{proposition}
A pivotal difference between $\mathbb{Q}_p$ and $\mathbb{R}$ is that in $\mathbb{Q}_p$ the digit expansion is carried out ``to the left'' and not ``to the right''. Moreover, the $p$-adic expansion \eqref{eq:padic_expansion} is \textbf{unique}, which is not the case with real numbers (\textit{e.g.}, $1.000... = 0.999...$ are two different expansions in base 10 of the same number). 

Addition and multiplication in $\mathbb{Z}_p$ can be performed using this $p$-adic expansion as we would normally do with $\mathbb{Z}$, except quantities carry over infinitely ``to the left''. The same holds in $\mathbb{Q}_p$ by accounting for the ``decimal point''. 

\begin{example}
    Take $p=2$. The numbers $-1 = \textcolor{red}{\cdots 111}$, $12 = \textcolor{red}{1100}$ and $\frac{1}{3} = \textcolor{red}{\cdots 10101011}$ are all $2$-adic integers. The numbers $\frac{1}{2} = \textcolor{red}{0.1}$ and $\frac{5}{4} = \textcolor{red}{1.01}$ are elements of $\mathbb{Q}_2$ but not $2$-adic integers. 
\end{example}

\begin{example}
Take $p=5$. $\frac{1}{2} = \textcolor{red}{\cdots 2223}$ and $\frac{1}{4} = \textcolor{red}{\cdots 3334}$ are both $5$-adic integers. 
$\frac{5}{2}$ appends a zero to the right of $\frac{1}{2}$, $\frac{5}{2} = \textcolor{red}{\cdots 2230}$.
Note that the digit-wise addition $\frac{1}{4} + \frac{1}{4} = \frac{1}{2}$ works (modulus 5) when we carry over to the left. 
\end{example}

Naturally, we have 
$\mathbb{Q} \subset \mathbb{Q}_p$, $\mathbb{Z} \subset \mathbb{Z}_p \subset \mathbb{Q}_p$, and $\mathbb{Z}_p \cap \mathbb{Q} = \{\frac{a}{b} \in \mathbb{Q} \,:\, p \nmid b\}$. 
Topologically, $\mathbb{Q}_p$ is very different from $\mathbb{R}$.%
\footnote{Formally, $\mathbb{Q}_p$ is a totally disconnected Hausdorff topological space. $\mathbb{Z}_p$ is compact and $\mathbb{Q}_p$ is locally compact. \citep[Corollaries 4.2.6-7]{gouvea2020padic}.}
 %
For example, unlike $\mathbb{R}$,  $\mathbb{Q}_p$ is not an ordered field, \textit{i.e.}, expressions such as $x \ge 0$ or $x \le y$ do not make sense when $x, y \in \mathbb{Q}_p$. 
This requires a new approach to define $p$-adic binary classifiers, as we shall see in \S\ref{sec:classification}. %



\subsection{Properties of $\mathbb{Q}_p$} 

\paragraph{Balls in $\mathbb{Q}_p$.} Let $a \in \mathbb{Q}_p$ and $r \in \mathbb{R}_+$. The (closed) ball with center $a$ and radius $r$, denoted $\bar{B}_{r}(a)$, is the set 
\begin{align}
    \bar{B}_{r}(a) := \{x \in \mathbb{Q}_p \,:\, |x-a|_p \le r \}.
\end{align}
Since the image of the $p$-adic absolute value function is the set $\{0\} \cup \{p^n \,:\, n \in \mathbb{Z}\}$, many balls with different radii are identical---for example, $\bar{B}_r(a) = \bar{B}_{p^n}(a) $ for any $r \in [p^n, p^{n+1}[$. 
Therefore, for most purposes it suffices to consider balls of the form $\bar{B}_{p^n}(a)$ where $n \in \mathbb{Z} \,\cup\, \{-\infty\}$. 
An intriguing property of $\mathbb{Q}_p$ (and more generally of ultrametric spaces) is that 
\textbf{every point in a ball is a center of that ball}---that is,  
$\bar{B}_r(a) = \bar{B}_r(x)$ for any $x \in \bar{B}_r(a)$. 
Another important property is that \textbf{any two balls are either disjoint or one is nested into the other}---that is, for any $a, b \in \mathbb{Q}_p$ and $r, s \in \mathbb{R}_+$ with $r \le s$, either $\bar{B}_r(a) \cap \bar{B}_s(b) = \varnothing$ or $\bar{B}_r(a) \subseteq \bar{B}_s(b)$. 
Topologically, balls in $\mathbb{Q}_p$ are simultaneously closed and open (``clopen''). 

\begin{example}
Any ball with zero radius is a singleton, $\bar{B}_0(a) = \{a\}$. 
\end{example}

\begin{example} 
The set of $p$-adic integers is a unit ball, $\mathbb{Z}_p = \bar{B}_1(0)$. The choice of $0$ as the center is arbitrary: we also have $\mathbb{Z}_p = \bar{B}_1(x)$ for any $x \in \mathbb{Z}_p$. 
We can decompose $\mathbb{Z}_p$ as the disjoint union of $p$ smaller balls:
\begin{align}\label{eq:Zp_partition}
\mathbb{Z}_p = \bar{B}_1(0) = \bar{B}_{\frac{1}{p}}(0) \cup \bar{B}_{\frac{1}{p}}(1) \cup \ldots \cup \bar{B}_{\frac{1}{p}}(p-1).
\end{align}
This shows that $\mathbb{Z}_p$ has a hierarchical structure, as illustrated in Figure~\ref{fig:Qp_hierarchy}.  Note that something similar happens with any ball of $\mathbb{Q}_p$ with radius $p^n$: it can be decomposed as a disjoint union of $p$ smaller balls, each with  radius $p^{n-1}$.  
Additional properties of $p$-adic balls 
are shown in Appendix~\ref{sec:padic_balls_properties}. 
\end{example}

\begin{figure}[t]
    \centering
    \includegraphics[width=0.8\columnwidth]{figures/Qp_hierarchy.pdf}%
    \caption{Hierarchical structure of $\mathbb{Q}_p$ for $p=2$. Balls in $\mathbb{Q}_p$ are either nested or disjoint and each ball is associated to a node in the hierarchy. Shown are two nested balls, $\bar{B}_1(0)$, which is the set of $p$-adic integers, and $\bar{B}_{1/4}(4)$, which is contained in the former and whose elements are $p$-adic integers of the form $\textcolor{red}{\cdots 10}$.}
    \label{fig:Qp_hierarchy}
\end{figure}


\begin{example}[$p$-adic balls are strings]\label{ex:padic_strings}
    Any ball $\bar{B}_r(a) \subseteq \mathbb{Z}_p$ with $r>0$ can be identified with a string over an alphabet with $p$ symbols. Namely, with $r = p^{-n}$ and $n \in \mathbb{N}$, any element of $\bar{B}_r(a)$ has a $p$-adic expansion of the form $$x = \textcolor{red}{* a_{n-1} \ldots a_0},$$ where $a_i$ is the $i$\textsuperscript{th} digit of $a$ and the wildcard $*$ denotes arbitrary symbols occurring to the left. Therefore, elements of $\mathbb{Z}_p$ may be regarded as ``infinite strings'' and balls of $\mathbb{Z}_p$ may be regarded as strings of finite length. 
\end{example}

In Appendix~\ref{sec:kraft}, we use the connection above between $p$-adic balls and strings to obtain a simple proof of  \textbf{Kraft's inequality} \citep{kraft1949device}, a key result in information theory which establishes a necessary and sufficient condition for a code to be a prefix code. (To the best of our knowledge, this proof based on $p$-adic balls is novel.)

\section{$p$-adic Classification}\label{sec:classification}

Let $d \in \mathbb{N}$ and denote by $\mathbb{Q}_p^d$ the $d$-dimensional vector space  over the field $\mathbb{Q}_p$ with the standard vector addition and scalar multiplication. 
In this section, we discuss $p$-adic binary classifiers, which take as input $d$-dimensional feature vectors $x \in \mathbb{Q}_p^d$ and predict outputs $y \in \{\pm 1\}$. 
We assume training data $\mathcal{D} = \left\{(\xx{i}, \yy{i})\right\}_{i=1}^n \subseteq \mathbb{Q}_p^d \times \{\pm 1\}$, with $n \in \mathbb{N}$. 
We consider both one-class classification (also known as anomaly detection), where the training set contains only positive examples, and two-class classification, where the training set contains both positive ($+1$) and negative examples ($-1$).

Binary classifiers operating over $\mathbb{R}$ predict according to the rule $\hat{y}(x) = \mathrm{sign}(f(x))$, where $f: \mathbb{R}^d \rightarrow \mathbb{R}$ is a discriminant function and $\mathrm{sign}(.)$ is the sign function. 
However, we cannot apply a similar prediction rule to $p$-adic classifiers with $f: \mathbb{Q}_p^d \rightarrow \mathbb{Q}_p$, since $\mathbb{Q}_p$ is not an ordered field, 
and therefore we cannot use a sign function as above. 
Instead, we define a classification rule where the prediction is $+1$ iff the argument is in $\mathbb{Z}_p$ and $-1$ otherwise: 
\begin{align}\label{eq:padic_decision_rule}
\hat{y}(x) := \left\{
\begin{array}{ll}
+1, & \text{if $|f(x)|_p 
\le 1$}\\
-1, & \text{otherwise.}
\end{array}
\right.
\,\,=\,\,
\left\{
\begin{array}{ll}
+1, & \text{if $f(x) \in \mathbb{Z}_p$}\\
-1, & \text{otherwise.}
\end{array}
\right.
\end{align}
The next subsections address the unidimensional case ($x \in \mathbb{Q}_p$) and then extend it to $d\ge 1$ $p$-adic features ($x \in \mathbb{Q}_p^d$). 

\subsection{Linear classifiers: Unidimensional case}\label{sec:linear_classifiers_unidimensional}

\begin{figure}[tp]
    \centering
    \includegraphics[width=0.65\columnwidth]{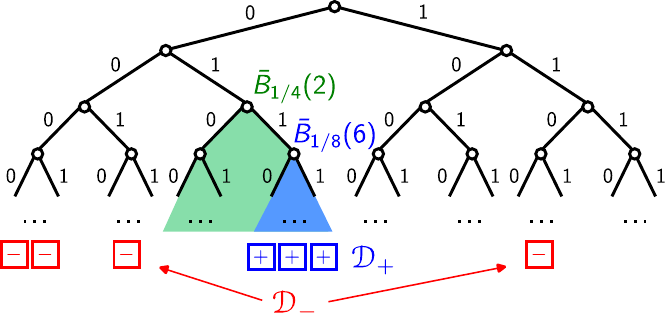}
    \includegraphics[width=0.33\columnwidth]{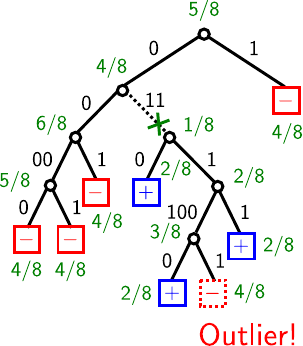}
    \caption{\textbf{Left:} A separable dataset in $\mathbb{Q}_2$ and enclosing balls. The positive examples are of the form \textcolor{red}{*110} and the negative examples either of the form \textcolor{red}{*00} or \textcolor{red}{*1}. We choose $x_i=6 = \textcolor{red}{110}$ as representative of the positive class. The point $x_j=-2 = \textcolor{red}{\cdots 1110}$ is maximally distant from $x_i$. Using Proposition~\ref{prop:binary_classifier_conditions}, we obtain $w^\star = \sfrac{1}{8}$ and $b^\star = -\sfrac{3}{4}$,  corresponding to the minimal enclosing ball $\bar{B}_{\sfrac{1}{8}}(6) = \textcolor{red}{*110}$. The maximal enclosing ball is $\bar{B}_{\sfrac{1}{4}}(6) = \textcolor{red}{*10}$. 
    \textbf{Right:} Adding the outlier $x_- = \textcolor{red}{\cdots 11001110}$ turns the dataset not separable. Let the dataset be $\mathcal{D}_+ = \{\textcolor{red}{*0110}, \textcolor{red}{*01001110}, \textcolor{red}{*11110}\}$ and $\mathcal{D}_- = \{\textcolor{red}{*00000}, \textcolor{red}{*10000}, \textcolor{red}{*100}, \textcolor{red}{*11001110}, \textcolor{red}{*1}\}$ (the left continuation is not important). We represent these points in a tree where each leaf represents the largest ball containing a single point and the nodes represent splitting points (edges are labeled with one or more symbols). 
    To determine the classifier with the smallest misclassification error, we consider a ball rooted at each node (or leaf) and compute the training error associated with that ball. In this example, the optimal classifier corresponds to either of the balls $\bar{B}_{\sfrac{1}{4}}(2) = \textcolor{red}{*10}$ or $\bar{B}_{\sfrac{1}{8}}(6) = \textcolor{red}{*110}$, where the dashed edge is cut, leading to a training error of 1/8.
    }
    \label{fig:linear_binary_classifier}
\end{figure}

In the linear and unidimensional case  ($x \in \mathbb{Q}_p$), we use \eqref{eq:padic_decision_rule} with  
$f(x) = wx + b$, where $w, b \in \mathbb{Q}_p$ are model parameters. 
If $w = 0$, this becomes a trivial always-positive ($b \in \mathbb{Z}_p$) or always-negative ($b \notin \mathbb{Z}_p$) classifier, so we focus on the case $w\ne 0$. 
Since $|wx+b|_p = |w|_p \left|x + \frac{b}{w}\right|_p$ the decision rule  \eqref{eq:padic_decision_rule} becomes
\begin{align}\label{eq:ball_decision_rule}
\hat{y}(x) = \left\{
\begin{array}{ll}
+1, & \text{if $x \in \bar{B}_r(a)$}\\
-1, & \text{otherwise,}
\end{array}
\right. 
\end{align}
where $a = -\frac{b}{w}$ and $r = \frac{1}{|w|_p}$. 
Therefore, a unidimensional $p$-adic binary linear classifier \textbf{simply checks if the input $x$ lies within a  $p$-adic ball}. 
To train such a classifier, we need to find a ball that encloses as many positive training examples as possible and  excludes most negative training examples. 
Perfect linear separation is possible iff there is a ball that perfectly separates one class from the other---this problem can be solved in $\mathcal{O}(n)$ time. 
In one-class classification problems, no negative examples are available, so a reasonable criterion is to search for the ball with the smallest radius which encloses the positive examples---which parallels similar objectives in Euclidean/Hilbert spaces over $\mathbb{R}$ \citep{nolan1991excess,scholkopf1999support}. For $p$-adic classifiers, this is equivalent to finding the least common ancestor node of the positive examples. 
A full characterization, proved in Appendix~\ref{sec:proof_binary_classifier_conditions}, is given below (and illustrated in Figure~\ref{fig:linear_binary_classifier}). 

\begin{proposition}\label{prop:binary_classifier_conditions}
    Let $\mathcal{D} = \mathcal{D}_+ \cup \mathcal{D}_-$ be a training set with positive/negative examples $\mathcal{D}_+ = \{\xx{1}, ..., \xx{m}\}$ and $\mathcal{D}_- = \{\xx{m+1}, ..., \xx{n}\}$. Suppose that $m\ge 2$ and that $\mathcal{D}$ is linearly separable, \textit{i.e.}, there is $(w,b)$ such that $|w x_+ + b|_p \le 1$ $\forall x_+ \in \mathcal{D}_+$ and $|w x_- + b|_p > 1$ $\forall x_- \in \mathcal{D}_-$. Then:
    \begin{enumerate}
        \item Pick any $\xx{i} \in \mathcal{D}_+$ and let $\xx{j} \in \arg\max_{x_+ \in \mathcal{D}_+} |x_+ - \xx{i}|_p$ be a maximally distant positive example. Then, $w^\star :={1}/({\xx{j}-\xx{i}})$ and $b^\star :=-{\xx{i}}/({\xx{j}-\xx{i}})$ parametrize a separating linear classifier. This classifier satisfies $w^\star \xx{i} + b^\star = 0$ and $w^\star \xx{j} + b^\star = 1$. 

        \item The classifier $(w^\star, b^\star)$ above corresponds to the enclosing ball $\bar{B}_r(a)$ with $a=\xx{i}$ and $r = |\xx{j} - \xx{i}|_p$---this is the (unique) minimal enclosing ball that contains $\mathcal{D}_+$. 
        \item Let $\xx{k} \in \arg\min_{x_- \in \mathcal{D}_-} |x_- - \xx{i}|_p$ be a maximally close negative example (and due to separability of $\mathcal{D}$ and ultrametricity, also maximally close to any point in $\mathcal{D}_+$). Then, $\bar{B}_{r'}(a)$ with $a=\xx{i}$ and $r'=p^{-1}|\xx{k} - \xx{i}|_p$ is the (unique) maximal enclosing ball that contains $\mathcal{D}_+$ and defines a separating linear classifier. 
        \item Any ball with center in $\xx{i}$ and radius in $[r, r']$ is an enclosing ball defining a separating linear classifier. Any separating linear classifier is of this form. 
    \end{enumerate}
\end{proposition}

If the problem is not separable, it is possible to find the best enclosing ball with respect to some loss function (\textit{e.g.}, misclassification rate) with a $\mathcal{O}(n)$ algorithm (see Figure~\ref{fig:linear_binary_classifier}): 
\begin{enumerate}
    \item First, build a tree whose leaves represent the points in $\mathcal{D}_+$ and whose nodes represent the balls containing subsets of these points (edges represent one or more digits, and only nodes with multiple children need to be considered).
    \item Then, examine each node and compute the loss value by counting how many positive and negative points are enclosed by the corresponding ball. Pick the best node.
\end{enumerate}

It is interesting to compare this procedure with what happens in the real line ($\mathbb{R}$) where this unidimensional problem is also tractable---one needs to examine what happens in the intervals delimited by consecutive points instead of reasoning about nodes. 



\subsection{Nonlinear classifiers: Unidimensional case}\label{sec:nonlinear_classifiers} 

We now consider the decision rule \eqref{eq:padic_decision_rule} with a nonlinear discriminant function $f: \mathbb{Q}_p \rightarrow \mathbb{Q}_p$, such as a polynomial of degree $s\ge 1$, 
$f(x) = c \prod_{j=1}^s (x-a_j)$ for $c, a_1, ..., a_s \in \mathbb{Q}_p$. In this case (assuming $c \ne 0$) the decision is $\hat{y}(x)=+1$ iff $\prod_{j=1}^s |x-a_j|_p \le r$, with $r = 1/|c|_p$. 
We show that in this case the positive region may be a \textit{union} of $p$-adic balls.
\footnote{Note that polynomial classifiers with sufficiently large degree can overfit the training data with perfect accuracy by setting $s := |\mathcal{D}_+|$, $a_j := \xx{j}$ for each $\xx{j} \in \mathcal{D}_+$, and choosing $r = 1/|c|_p <  \min_{x_- \in \mathcal{D}_-} \prod_{x_+ \in \mathcal{D}_+} |x_- - x_+|_p$. This ensures $f(x_+) = 0$ for any $x_+ \in \mathcal{D}_+$ and $|f(x_-)|_p > 1$ for any $x_+ \in \mathcal{D}_+$.} %

\begin{proposition}[2\textsuperscript{nd} order $p$-adic classifier]\label{prop:second_order} 
Let $f(x) = c(x-a_1)(x-a_2)$, with $\hat{y}(x)$ defined as \eqref{eq:padic_decision_rule}. 
Let $|a_1-a_2|_p = p^{-k_{12}}$ and $r = 1/|c|_p = p^{-k}$ for some $k_{12}, k \in \mathbb{Z}$. 
Then
\begin{enumerate}
    \item If $k > 2k_{12}$, $\hat{y}(x) = +1$ iff 
        $x \in \bar{B}_{p^{-k+k_{12}}}(a_1) \cup \bar{B}_{p^{-k+k_{12}}}(a_2)$.
    \textit{i.e.}, the positive region is the disjoint union of two balls with the same radius.
    \item If $k \le 2k_{12}$, $\hat{y}(x) = +1$ iff $x \in \bar{B}_{p^{-\lceil k/2\rceil}}(a_1) = \bar{B}_{p^{-\lceil k/2\rceil}}(a_2)$, \textit{i.e.}, the positive region is a single ball and the classifier is equivalent to a linear classifier.
\end{enumerate}
\end{proposition}

\begin{figure}
    \centering
    \includegraphics[width=0.64\columnwidth]{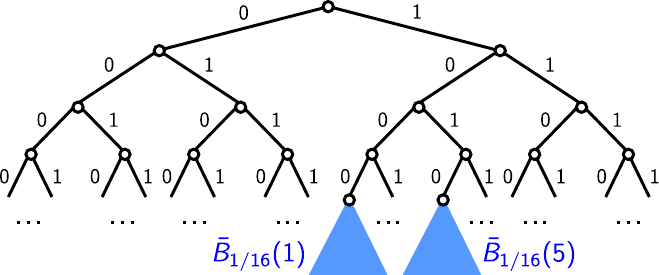}
    \caption{Example of a 2\textsuperscript{nd} order classifier with $f(x) = \frac{1}{64}(x-1)(x-5)$. The positive region is the union $\bar{B}_{\frac{1}{16}}(1) \cup \bar{B}_{\frac{1}{16}}(5)$.}
    \label{fig:second_order_classifier}
\end{figure}

The proof is in Appendix~\ref{sec:proof_second_order}.%
\footnote{For $s>2$ we 
have the union of at most $s$ balls, 
but the balls do not need to have all the same size.}

\begin{example}
Consider the classifier in $\mathbb{Q}_2$ defined by $|f(x)|_p = |\frac{1}{64}(x^2 -6x +5)|_p \le 1$. The function decomposes as $f(x) = \frac{1}{64}(x-1)(x-5)$. We have $k=6$ and $k_{12}=2$. Since $k>2k_{12}$ and we have $-k+k_{12} = -4$, the positive region is defined by $\bar{B}_{\frac{1}{16}}(1) \cup \bar{B}_{\frac{1}{16}}(5)$, which consists of numbers of the form $\textcolor{red}{*0001}$ or $\textcolor{red}{*0101}$. See Figure~\ref{fig:second_order_classifier}. 
\end{example}

\subsection{Linear classifiers: Multidimensional inputs} 

Let now $d\ge 1$. The input space is now a $d$-dimensional vector space over $\mathbb{Q}_p$, which we denote by $\mathbb{Q}_p^d$. 
We extend the framework of \S\ref{sec:linear_classifiers_unidimensional} to this scenario as follows.%
\footnote{An alternative way to extend \S\ref{sec:linear_classifiers_unidimensional} which deserves consideration would be to construct a ball-like decision rule similar to \eqref{eq:ball_decision_rule} by defining balls in $\mathbb{Q}_p^d$, which can be done by defining a norm on $\mathbb{Q}_p^d$. An appealing choice is the supremum norm, $\|x\|_p := \max_{1\le i\le d} |x_i|_p$, which endows $\mathbb{Q}_p^d$ with the structure of a ultrametric space, keeping the same non-Archimedean flavor as $\mathbb{Q}_p$. 
Unfortunately, classifiers using the rule $\|x-a\|_p = \max_i |x_i-a_i|_p \le r$ seem quite restrictive---they require $x_i \in \bar{B}_r(a_i)$ for each $i$, so they are similar to a conjunction of binary classifiers applied to each feature, all constrained to have the same radius. We therefore opted by following the construction presented in this section based on linear sums.} %

We assume $\hat{y}(\cdot; w, b)$ is defined as in \eqref{eq:padic_decision_rule} with 
$f(x) = w^\top x + b$, 
where $w \in \mathbb{Q}_p^d$, $b \in \mathbb{Q}_p$, and $w^\top x := \sum_{i=1}^d w_{i} x_{i}$. %
Let $\mathcal{H}_{\mathbb{Q}_p^d} := \{\pm\hat{y}(\cdot; w, b) : w \in \mathbb{Q}_p^d, b \in \mathbb{Q}_p\}$ be the hypothesis class of \textbf{$\mathbb{Q}_p$-linear classifiers}, 
and $\mathcal{H}_{\mathbb{R}^d} := \{\hat{y}_\mathbb{R}(\cdot; w, b) : w \in \mathbb{R}^d, b \in \mathbb{R}\}$ be the hypothesis class of \textbf{$\mathbb{R}$-linear classifiers}, where  $\hat{y}_\mathbb{R}(x; w, b) = \mathrm{sign}(w^\top x + b)$. 

The next result, proved in Appendix~\ref{sec:proof_classifier_properties}, shows that some problems which cannot be solved by real linear classifiers---such as XOR or parity problems, which \citet{minsky1988perceptrons} have shown cannot be solved by ``finite order perceptrons'', or counting problems, which require ``second-order perceptrons''---are easy for $p$-adic linear classifiers, and vice-versa---count thresholding problems are trivial for real linear classifiers but cannot be solved directly by $p$-adic linear classifiers. 

\begin{proposition}[Properties of $\mathbb{Q}_p$-linear classifiers]\label{prop:properties_padic_classifiers} 
\begin{enumerate} 
    \item[]
    \item For $d=2$ and any prime $p$, classifiers in $\mathcal{H}_{\mathbb{Q}_p^d}$ can compute any Boolean function. This includes XOR (which $\mathcal{H}_{\mathbb{R}^d}$ cannot solve). 
    \item For $d\ge 2$ and any prime $p$,  classifiers in $\mathcal{H}_{\mathbb{Q}_p^d}$ can solve congruence problems modulo $p^n$ (which include parity checks) and counting problems---which classifiers in $\mathcal{H}_{\mathbb{R}^d}$ cannot solve---but they cannot solve count thresholding problems, which are solvable by $\mathbb{R}$-linear classifiers.
\end{enumerate}
\end{proposition}

Finally, the next proposition, proved in Appendix~\ref{sec:proof_classifier_conditions}, generalizes our previous result for the unidimensional case  (Proposition~\ref{prop:binary_classifier_conditions}). 

\begin{proposition}\label{prop:classifier_conditions} 
Assume $\hat{y}(\cdot; w,b) \in \mathcal{H}_{\mathbb{Q}_p^d}$ classifies correctly all points in  $\mathcal{D}_+ := \{\xx{1},..., \xx{m}\} \subseteq \mathbb{Q}_p^d$ and denote by $R(w, b) := \{x \in \mathbb{Q}_p^d \,:\, |w^\top x + b|_p \le 1\}$  the ``positive region'' according to this classifier. Then:
\begin{enumerate}
    \item For any $i \in [m]$, the parameters $(w,b')$ with $b'=-w^\top \xx{i}$ define an identical classifier, \textit{i.e.}, we have $R(w, b) = R(w, b')$. In other words, 
    $R(w, b) = \{x \in \mathbb{Q}_p^d \,:\, |w^\top (x - \xx{i})|_p \le 1\}$ for arbitrary choice of $i \in [m]$. 
    \item If the classifier is ``tight'' in $\mathcal{D}_+$ (meaning that $\exists j \in [m]$ such that $|w^\top \xx{j} + b|_p = 1$) then there is a parametrization $(w^\star, b^\star) \in \mathbb{Q}_p^{d} \times \mathbb{Q}_p$ satisfying ${w^\star}^\top \xx{i} + b^\star = 0$ and ${w^\star}^\top \xx{j} + b^\star = 1$ for some $i,j \in [m]$ such that $R(w, b) = R(w^\star, b^\star)$, \textit{i.e.}, $\hat{y}(\cdot; w^\star,b^\star) = \hat{y}(\cdot; w,b)$. 
\end{enumerate}
\end{proposition}

The second item in Proposition~\ref{prop:classifier_conditions} requires as a condition that the classifier is tight. Note, however,  that this is really not an additional condition---for any classifier which classifies correctly all points in $\mathcal{D}_+$, we can always multiply $w$ and $b$ by some scalar $\lambda$ with $|\lambda|_p \ge 1$ such that the classifier becomes tight. 
Note also that this property is not analogous to anything similar for the real numbers, not even for $d=1$.

\subsection{Learning a  $p$-adic linear classifier}\label{sec:learning_padic_classifier}

The previous section established what can and cannot be learned by a $d$-dimensional $\mathbb{Q}_p$-linear classifier.  
We now describe an approximate algorithm to learn a classifier from data. 
In the next two remarks we define, for $x \in \mathbb{Q}_p^d$, the $p$-adic norm $\|x\|_p := \max_{i \in [d]} |x_{i}|_p$. 
This is a valid norm in the vector space $\mathbb{Q}_p^d$ and, defining $d(x, z) := \|x-z\|_p$, it turns $\mathbb{Q}_p^d$ into an ultrametric space.  

\begin{remark}[Reduction to integers]\label{rem:reduction_integers}
Let $\mathcal{D} = \mathcal{D}_+ \cup \mathcal{D}_- = \{\xx{1}, ..., \xx{n}\}$ be the training inputs. 
We can assume all inputs are $p$-adic integers---$\xx{j} \in \mathbb{Z}_p^d$ for all $j \in [n]$---without loss of generality. 
To see why, assume $(w, b) \in \mathbb{Q}_p^{d+1}$;  let $i$ correspond to the example where $\|\xx{i}\|_p$ is largest and let $p^{m}$ denote this quantity. 
Then, define $\xx{j}' := p^m \xx{j} \in \mathbb{Z}_p^d$ for all $j \in [n]$, and let $w' := p^{-m} w$. 
We have that $|{w'}^\top \xx{j}' + b|_p = |{w}^\top \xx{j} + b|_p$. Therefore, we have an equivalent classification problem where all data consist of $p$-adic integers, and the weight vector $w$ is scaled. 
\end{remark}


\begin{remark}[Integer weights]
There is also a formulation equivalent to \eqref{eq:padic_decision_rule} which ensures $(w, b) \in \mathbb{Z}_p^{d+1}$, so that the learning can be reduced to search over the $p$-adic integers: find $(w', b') \in \mathbb{Z}_p^{d+1}$ and $\ell \in \mathbb{Z}$ such that $|w'^\top x_+ + b'|_p \le p^{-\ell}$ for all positive examples $x_+$ and $|w'^\top x_- + b'|_p > p^{-\ell}$ for all negative examples $x_-$. 
Then we have that $w = p^{-\ell} w'$ and  $b = p^{-\ell} b'$ lead to a separating classifier. 
\end{remark}

\begin{algorithm}[t]
   \caption{Training a $p$-adic classifier with beam search.}
   \label{alg:padic_beam_search}
   \small
\begin{algorithmic}[1]
    \State {\bfseries Input:} dataset $\mathcal{D} := \mathcal{D}_+ \cup \mathcal{D}_- \subseteq \mathbb{Z}_p^{d+1}$, beam size $k$, maximum number of mistakes $\varepsilon_{\max}$, maximum depth $\delta_{\max}$
    \State $\delta := 0$ \Comment{current depth}
    \State Create node $n$ with $n.w := 0$ \Comment{root node}
    \State $\mathcal{A} := \{n\}$ \Comment{active nodes in the beam}
    \Repeat
    \State $\mathcal{N} := \varnothing$ \Comment{admissible nodes}
    \For{$n \in \mathcal{A}$}
        \For{$a \in \{0,...,p-1\}^{d+1}$} \Comment{expand node}
            \State $w := n.w + a p^{\delta}$
            \State \Comment{count positive and negative errors}
            \State $\varepsilon_+, \varepsilon_- := \textsc{Evaluate}(w/p^{\delta+1}, \mathcal{D})$ 
            \If{$\varepsilon_+ \le \varepsilon_{\max}$}
            \State Create node $c$ with $c.w := w$
            \State $\mathcal{N} := \mathcal{N} \cup \{c\}$
            \EndIf
        \EndFor
    \EndFor
    \State $\delta := \delta + 1$
    \State \Comment{keep the $k$ nodes with fewest mistakes $\varepsilon_+ + \varepsilon_-$}
    \State $\mathcal{A} := \textsc{Best-$k$}(\mathcal{N})$ 
    \If{there is $n \in \mathcal{A}$ with $\le \varepsilon_{\max}$ mistakes}
        \State \Return $w := n.w / p^{\delta}$
    \EndIf
    \Until{$\delta = \delta_{\max}$}
\end{algorithmic}
\end{algorithm}

Without loss of generality, we assume that inputs are $(d+1)$-dimensional and have a constant feature, $x_{d+1}=1$, so that we can drop the bias parameter. 
Algorithm~\ref{alg:padic_beam_search} shows a simple beam search algorithm for training a linear $p$-adic classifier. 
The algorithm assumes that all inputs $\xx{i}$ are in $\mathbb{Z}_p^{d+1}$, which can be ensured with the preprocessing in Remark~\ref{rem:reduction_integers}. 
We denote the number of positive and negative errors as 
$\varepsilon_+ := |\{\xx{i} \in \mathcal{D}_+ \,:\, \hat{y}(\xx{i}; w) = -1\}|$ and $\varepsilon_- := |\{\xx{i} \in \mathcal{D}_- \,:\, \hat{y}(\xx{i}; w) = +1\}|$, respectively. 
Each node $n$ with depth $\delta$ in the search tree has associated a weight vector $n.w \in \mathbb{Z}_p^{d+1}$ where only the $\delta$ most significant bits matter. For any descendant node, the number of positive errors $\varepsilon_+$ can only increase and the number of negative errors $\varepsilon_-$ can only decrease. Therefore, $\varepsilon_+$ is a lower bound for the number of mistakes achieved by any descendant node. 
The worst case runtime complexity of Algorithm~\ref{alg:padic_beam_search}  is $\mathcal{O}(k \delta_{\max} n p^{d+1})$, which is linear on the dataset size but grows exponentially fast with the number of features $d$. 

In \S\ref{sec:representation_learning}, we experiment with this algorithm to learn a $\mathbb{Q}_p$-linear classifier for solving logical inference problems. 

\section{$p$-adic Regression}\label{sec:regression}

In linear regression, 
 the goal is to estimate a target $y \in \mathbb{Q}_p$ with  $\hat{y}(x) = w^\top x + b$. Given the relation between $p$-adic numbers and strings (Example~\ref{ex:padic_strings}), we may informally think of $p$-adic regression as a sequence-to-sequence problem. 
Given a dataset $\mathcal{D} = \{(\xx{i}, \yy{i})\}_{i=1}^n$, we formulate the problem as that of finding $w$ and $b$ that minimize the sup-norm of the residuals, 
$\max_{i \in [n]} |w^\top \xx{i} + b - \yy{i}|_p$. 

\subsection{Linear regression: Unidimensional case}\label{sec:linear_regression_unidimensional} 

Consider first the unidimensional case  $x \in \mathbb{Q}_p$. 
The following result shows that this problem can be solved efficiently. 

\begin{proposition}\label{prop:unidimensional_regression}
    Assume that $\xx{i} \ne \xx{j}$ for any $i\ne j$. Then, 
    the following algorithm finds   $(w^\star, b^\star) \in \arg\min_{w,b} \max_{i \in [n]} |w\xx{i} + b - \yy{i}|_p$ in time $\mathcal{O}(n^2)$: 
\begin{enumerate}
    \item Choose $k \in [n]$ arbitrarily. 
    \item Find $j^\star \in \arg\min_{j\ne k} \max_{i\ne k} |\xx{i}-\xx{k}|_p \left| \frac{\yy{j}-\yy{k}}{\xx{j}-\xx{k}} - \frac{\yy{i}-\yy{k}}{\xx{i}-\xx{k}}\right|_p$ in $\mathcal{O}(n^2)$ time.
    \item Set $w^\star = \frac{\yy{j^\star}-\yy{k}}{\xx{j^\star}-\xx{k}}$ and $b^\star = \yy{k} - w^\star \xx{k}$. (Note that we also have $b^\star = \yy{j} - w^\star \xx{j}$.)
\end{enumerate}
\end{proposition}

The proof is in Appendix~\ref{sec:proof_prop_unidimensional_regression}. 
Note that the solution ($w^\star$, $b^\star$) above satisfies $w^\star \xx{j} + b = \yy{j}$ and $w^\star \xx{k} + b = \yy{k}$, \textit{i.e.}, \textbf{there is an optimal solution that passes through the points $j$ and $k$}. This is very different from linear regression with the real numbers, and is a consequence of the ultrametric property of the $p$-adics. 
We will next see that this property holds also for the multidimensional case ($d\ge 1$). 

\subsection{Linear regression: Multidimensional case}\label{sec:linear_regression_multidimensional}

Assume now $d\ge 1$ and the overdetermined case $n \ge d+1$. 
As in \S\ref{sec:learning_padic_classifier}, we can eliminate the bias parameter by appending a constant feature $x_{d+1}^{(i)}=1$. 
By defining the design matrix $X \in \mathbb{Q}_p^{n \times (d+1)}$, the sup-norm can be written as $\|Xw - y\|_p$, which we want to minimize with respect to $w \in \mathbb{Q}_p^{d+1}$. 

The next result, proved in Appendix~\ref{sec:proof_regression_exemplar}, shows that, under a suitable invertibility condition,%
\footnote{A matrix $A \in \mathbb{Q}_p^{d \times d}$ is invertible if there is $A^{-1} \in \mathbb{Q}_p^{d \times d}$ such that $AA^{-1} = A^{-1}A = I_d$ (the identity).} %
there is an optimal solution which is exemplar-based, as in  \S\ref{sec:linear_regression_unidimensional}. 

\begin{proposition}\label{prop:regression_exemplar}
    Let $X \in \mathbb{Q}_p^{n \times (d+1)}$ be the design matrix with $n \ge d+1$, and assume that all its $(d+1)$-by-$(d+1)$ submatrices are invertible. 
    Assume also that the original design matrix $\tilde{X} \in \mathbb{Q}_p^{n \times d}$ before bias augmentation (\textit{i.e.}, $\tilde{X} := X_{1:n, 1:d}$) also has all its $d$-by-$d$ submatrices invertible. 
    Then, there is  an optimal solution $w^\star \in \arg\min_w \|Xw - y\|_p$ passing through $d+1$ points. 
\end{proposition}

It should be noted that the conditions of the proposition forbid us (among other things) to have $\xxtilde{i} = 0$ as an input, as this would lead to non-invertible submatrices. 
A consequence of Proposition~\ref{prop:regression_exemplar} is that, under the stated assumptions, we can find an optimal $w^\star$ by selecting all possible combinations of $d+1$ out of the $n$ training examples, for each such combination solve a linear $p$-adic system 
(which is guaranteed to have a solution since we assumed that all $(d+1)$-by-$(d+1)$ submatrices of $X$ are invertible) and then pick the combination whose optimal weight $w^\star$ lead to the smallest sup-norm. 
This algorithm runs in time $\mathcal{O}(n^{d+1})$.

\section{$p$-adic Representations}\label{sec:representation_learning}

We now look at representations (``embeddings'') in $\mathbb{Q}_p^d$. 
One reason why neural networks are so effective comes from their ability to learn internal representations, by mapping examples to points in $\mathbb{R}^d$. 
What happens in the $p$-adic space? 


Consider first the unidimensional case $\mathbb{Q}_p$ (a single ``embedding dimension'') and let us reason about the embedding of $n$ input examples $\{\xx{i}\}_{i=1}^n$ in this space. 
Since $\mathbb{Q}_p$ has a hierarchical structure (Figure~\ref{fig:Qp_hierarchy}), 
we can 
associate to each $\xx{i}$ the largest ball $B_{r_i}(\xx{i})$ that contains this example and no other examples. 
The different balls are arranged hierarchically and can be represented as a finite tree whose leaves correspond to each of the $n$ input examples. 
For example, if $p=2$, we obtain a binary tree where each ball $B_{r_i}(\xx{i})$ corresponds to a bit string---this is similar to \textbf{Brown clusters} \citep{brown1992class}, a popular representation technique in natural language processing. 
Therefore, \textit{we can think of 
Brown clusters as $p$-adic representations which contrast with continuous, real-vector representations.} 

\begin{figure*}[t]
    \centering
\includegraphics[width=0.72\columnwidth]{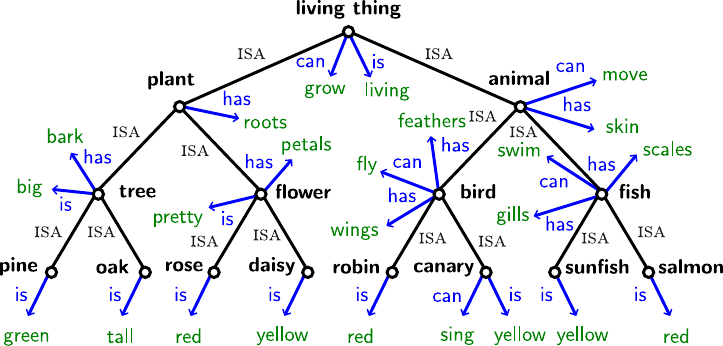}
\includegraphics[width=1\columnwidth]{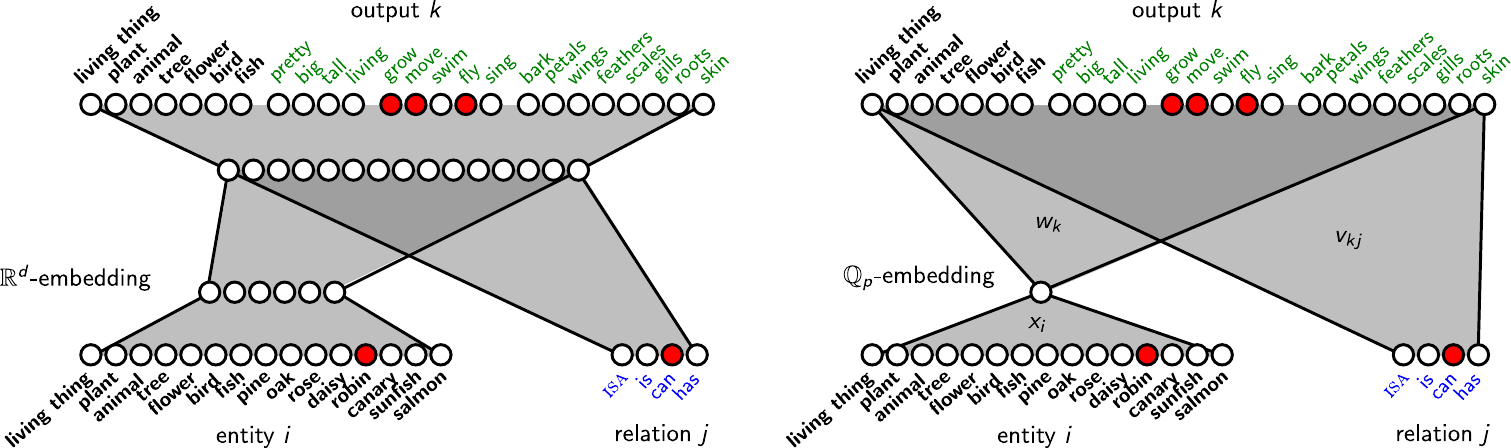}
    \caption{\textbf{Top:} Semantic network adapted from \citet{quillian1968semantic}. \textbf{Bottom left:} Neural network developed by \citet{rumelhart1990brain} and  \citet{mcclelland1995there} to answer queries using this semantic model. Given the active inputs ``robin can'', the network produces the completion ``grow move fly''. Note the two (non-linear) hidden layers, the first of which embeds input entities onto $\mathbb{R}^6$. \textbf{Bottom right:} A linear $p$-adic network with a single embedding dimension ($\mathbb{Q}_p$) which solves the same problem (colors and leaves attributes are excluded for simplicity). 
    }
    \label{fig:semantic_network}
\end{figure*}

\paragraph{Quillian's semantic networks.} 
In the previous example, all concepts correspond to leaves in a tree. It is appealing to think about internal nodes higher up in the tree as representing ``more general concepts'' associated to larger balls in $\mathbb{Q}_p$. These larger balls  enclose smaller balls (more specific concepts) forming a nested structure. 
Consider as an example the semantic network of Figure~\ref{fig:semantic_network} (top) corresponding to a simple hierarchical propositional model \citep{quillian1968semantic}. 

\citet{rumelhart1990brain} and \citet{mcclelland1995there} built a simple neural network with two hidden layers (Figure~\ref{fig:semantic_network}, bottom left) which is able to answer queries associated with this semantic network, through the embedding of concepts in $\mathbb{R}^6$.  
We next construct a more compact \textbf{linear} classifier with $p$-adic representations (with $p=2$) that encodes all the propositions of this semantic network (Figure~\ref{fig:semantic_network}, bottom right). Our $p$-adic classifier has a similar structure as the network in \citet{rumelhart1990brain} but only a single hidden layer (to encode the $p$-adic representations of concepts) instead of two, and it does not have any non-linear activations. It is a composition of linear functions and therefore it is still a linear classifier. The advantage of including the hidden representation layer is threefold: (i) it makes explicit the representation of the concepts, (ii) it reduces dimensionality, and (iii) it allows using a shared representation space to perform multiple tasks. 

We first show the representations and weights of the network without the ``is   green/red/yellow'' attribute in the leaves---in this case, a unidimensional $p$-adic representation space $\mathbb{Q}_p$ turns out to be sufficient. 
Then, we extend the network to include the ``is green/red/yellow'' attribute by using a two-dimensional $p$-adic representation space ($\mathbb{Q}_p^2$). Finally, we consider a new attribute, ``has leaves'', which defies the hierarchical structure of the semantic network: all plants have leaves \textit{except} the pine (which has needles instead of leaves). This exception can  be accommodated by using a three-dimensional $p$-adic representation space ($\mathbb{Q}_p^3$). 
In this construction, we always use $p=2$. 

\begin{table}[t]
    \caption{Representations of entities/concepts associated to the semantic network of Figure~\ref{fig:semantic_network}. Without the color attributes, only one feature dimension is needed (column $x$). With color attributes, we need a second feature dimension (column $x'$).}
    \label{tab:quillian_padic_entities}
    \vskip 0.15in
    \centering
    \small
    \begin{tabular}{crrrr}
\toprule
Entity & $x$ & $x$ (digits) & $x'$ & $x'$ (digits)\\
\midrule
living thing & 1 & \textcolor{red}{1} & 0 & \textcolor{red}{0}\\
plant & 4 & \textcolor{red}{100} & 0 & \textcolor{red}{0} \\
animal & 6 & \textcolor{red}{110} & 0 & \textcolor{red}{0}\\
tree & 16 & \textcolor{red}{10000} & 0 & \textcolor{red}{0}\\
flower & 24 & \textcolor{red}{11000} & 0 & \textcolor{red}{0}\\
bird & 18 & \textcolor{red}{10010} & 0 & \textcolor{red}{0}\\
fish & 26 & \textcolor{red}{11010} & 0 & \textcolor{red}{0}\\
pine & 64 & \textcolor{red}{1000000} & 1 & \textcolor{red}{1}\\
oak & 96 & \textcolor{red}{1100000} & 0 & \textcolor{red}{0}\\
rose & 72 & \textcolor{red}{1001000} & 2 & \textcolor{red}{10}\\
daisy & 104 & \textcolor{red}{1101000} & 4 & \textcolor{red}{100}\\
robin & 66 & \textcolor{red}{1000010} & 2 & \textcolor{red}{10}\\
canary & 98 & \textcolor{red}{1100010} & 4 & \textcolor{red}{100}\\
sunfish & 74 & \textcolor{red}{1001010} & 4 & \textcolor{red}{100}\\
salmon & 106 & \textcolor{red}{1101010} & 2 & \textcolor{red}{10}\\
\bottomrule
    \end{tabular}
\end{table}

\paragraph{Attributes excluding color.}
We first construct the model ignoring the color attributes (``is   green/red/yellow''). By looking at Figure~\ref{fig:semantic_network} (top) we can see that, with the exception of those attributes, all others are associated to a \textit{single} node of the semantic tree. 
We also see that each attribute is associated only to a single relation (\textit{e.g.}, ``grow'' is associated to \texttt{can} but not to \texttt{ISA}, \texttt{is}, or \texttt{has}). 
Our strategy is to (i) place the entity/concept representations in a $p$-adic tree with the same structure as the semantic network in  Figure~\ref{fig:semantic_network} and (ii) use a binary linear classifier for each attribute associated to a specific node of the tree.

Our network has the form in Figure~\ref{fig:semantic_network} (bottom right). 
We have two sets of inputs: 15 entities/concepts (living thing, plant, animal, ..., sunfish, salmon) and 4 relations (\texttt{ISA}, \texttt{is}, \texttt{can}, \texttt{has}). 
Like \citet{rumelhart1990brain}, we use one-hot encodings for both entities/concepts and relations. 
Entities/concepts are then fed into an embedding layer in $\mathbb{Q}_p$ with no bias parameter. This effectively means that they receive embeddings $x_1, \ldots, x_{15}$, where each $x_i \in \mathbb{Q}_p$. 
Relations $j$ use one-hot encodings directly. 
The representations of entities/concepts and relations are concatenated; 
therefore, when entity $i$ and relation $j$ are present in the input, this leads to a 5-dimensional vector $$[x_i, \underbrace{0, \ldots, 1, \ldots, 0}_{\text{one-hot for relation $j$}}] \in \mathbb{Q}_p^5.$$ 
These representations are shared across all attributes, and each attribute $k$ is associated to a binary classifier with 5-dimensional weights $[w_k, v_{k1}, \ldots, v_{k4}]$ and a bias parameter, which we set to zero. 
The output layer predicts 
\begin{align}\label{eq:quillian_padic}
    \hat{y}_k(i, j) &= \left\{
    \begin{array}{ll}
         +1 & \text{if $|w_{k} x_i + v_{kj}|_p \le 1$} \\
         -1 & \text{otherwise,}
    \end{array}
    \right.
\end{align} 
where (see \eqref{eq:ball_decision_rule}) the top condition is equivalent to $x_i \in \bar{B}_{1/{|w_k|_p}}\left(-{v_{kj}}/{w_k}\right)$. 
By examining the structure of the semantic network, it is straightforward to obtain parameters
$x_i$, $w_k$, and $v_{kj}$ that lead to the intended classifiers for each attribute $k$. We start by choosing  representations $x_i \in \mathbb{Q}_p$ that are compatible with the entities/concepts in  Figure~\ref{fig:semantic_network} (bottom left). The representations in Table~\ref{tab:quillian_padic_entities} satisfy this requirement. 
Then, for any attribute $k$, we consider the (unique) relation $j$ associated with that attribute. We first pick $w_k$ such that $1/|w_k|_p$ is at the correct level of the tree, following the right hand side of \eqref{eq:quillian_padic}---\textit{e.g.}, the attribute-relation pair ``ISA bird'' should encompass the entities/concepts bird, robin, and canary, whose common prefix is \textcolor{red}{$\cdots$0010}, therefore we need $1/|w_k|_p = 2^{-4}$ which is satisfied by $w_k = 2^{-4} = \sfrac{1}{16}$. Then, we pick $v_{kj}$ such that $-\frac{v_{kj}}{w_k}$ is a center of the desired ball---for example, for the attribute-relation pair ``ISA bird'' any center prefixed by \textcolor{red}{$\cdots$0010} will do. A possible such center is 2, which leads to $v_{kj} = -2w_k = -2\times 2^{-4} = -\sfrac{1}{8}$. 
Finally, for attribute-relation pairs that are incompatible we choose a ball disjoint from all the entity/concepts. 
This leads to the parameters shown in Table~\ref{tab:quillian_padic_relations}. 
This choice of parameters ensure correct classification for all choice of entities/concepts and relations in the input. 

\paragraph{Adding colors.} The strategy described above fails when we add color attributes, since the red and yellow color attributes are linked to multiple nodes (red is linked to rose, robin, salmon, and yellow is linked to daisy, canary, sunfish). This can be solved by adding a second embedding dimension $x_i'$ corresponding to the ``color'' feature (rightmost columns of Table~\ref{tab:quillian_padic_entities}). This requires organizing the entities/concepts according to a different hierarchy associated with the colors---fortunately, multidimensional $\mathbb{Q}_p$-linear classifiers allow representing multiple hierarchies. 
Now the input representations will be 6-dimensional vectors 
$[x_i, x_i', {0, ..., 1, ..., 0}] \in \mathbb{Q}_p^6$
and each attribute $k$ is associated to a binary classifier with 6-dimensional weights $[w_k, w_k', v_{k1}, \ldots, v_{k4}]$ and a zero bias. 
If the attribute $k$ is not a color, we set $w_k'=0$ and the classification rule is exactly as in \eqref{eq:quillian_padic}. 
If the attribute $k$ is a color (green, red, or yellow), we set $w_k=0$, which leads to  
\begin{align}\label{eq:quillian_colors_padic}
    \hat{y}_k(i, j) &= \left\{
    \begin{array}{ll}
         +1 & \text{if $|w_{k}' x_i' + v_{kj}|_p \le 1$} \\
         -1 & \text{otherwise.}
    \end{array}
    \right. 
\end{align}
The same logic as above works here, leading to the weights $w'_{k}$ and $v_{kj}$  in Table~\ref{tab:quillian_padic_relations}. 


\begin{table}[!t]
    \caption{Weights $w_k$ and $v_{kj}$ for each relation $j$ and attribute $k$. Each relation/attribute pair corresponds to a $2$-adic ball with radius $1/|w_k|$ and with $-v_{kj}/w_k$ as a center. For example, ``ISA bird'' corresponds to the ball $B_{2^{-4}}(2)$ which corresponds to $p$-adic digit expansions \textcolor{red}{$\cdots$0010}, containing the representations of bird, robin, and canary (see Table~\ref{tab:quillian_padic_entities}). Attributes $k$ which are not compatible with a relation $j$ (\textit{e.g.} ``ISA grow'') have weights $v_{kj} = -\frac{w_k}{2}$ (not shown in the table). Therefore they correspond to balls with center $\frac{1}{2}$ and radius $1/|w_k|$; since $1/|w_k| \le 1$ for all attributes $k$ in the table, these balls will contain only entities outside $\mathbb{Z}_2$ and therefore none of the entities in Table~\ref{tab:quillian_padic_entities}. 
    Color attributes are denoted in \textcolor{blue}{blue}, and for those we have $w_k = 0$ and the shown value is $w_k'$. For non-color attributes we have the opposite, $w_k' = 0$ and the shown value is $w_k$.}
    \label{tab:quillian_padic_relations}
    \vskip 0.15in
    \centering
    \small
\begin{tabular}{ccrrr}
\toprule
Relation $j$ & Attribute $k$ & $w_{k}$ (or \textcolor{blue}{$w_{k}'$}) & $v_{kj}$ & \\
\midrule
\texttt{ISA} & living thing & 1 & 0 & \\
\texttt{ISA} & plant & $\sfrac{1}{4}$ & 0 & \\
\texttt{ISA} & animal & $\sfrac{1}{4}$ & $\sfrac{-1}{2}$ & \\
\texttt{ISA} & tree & $\sfrac{1}{16}$ & 0 & \\
\texttt{ISA} & flower & $\sfrac{1}{16}$ & $\sfrac{-1}{2}$ & \\
\texttt{ISA} & bird & $\sfrac{1}{16}$ & $\sfrac{-1}{8}$ & \\
\texttt{ISA} & fish & $\sfrac{1}{16}$ & $\sfrac{-5}{8}$ & \\
\texttt{is} & pretty & $\sfrac{1}{16}$ & $\sfrac{-1}{2}$ & \\
\texttt{is} & big & $\sfrac{1}{16}$ & 0 & \\
\texttt{is} & living & 1 & 0 & \\
\textcolor{blue}{\texttt{is}} & \textcolor{blue}{green} & \textcolor{blue}{$\sfrac{1}{2}$} & \textcolor{blue}{$\sfrac{-1}{2}$} & \\
\textcolor{blue}{\texttt{is}} & \textcolor{blue}{red} & \textcolor{blue}{$\sfrac{1}{4}$} & \textcolor{blue}{$\sfrac{-1}{2}$} & \\
\textcolor{blue}{\texttt{is}} & \textcolor{blue}{yellow} & \textcolor{blue}{$\sfrac{1}{8}$} & \textcolor{blue}{$\sfrac{-1}{2}$} & \\
\texttt{is} & tall & $\sfrac{1}{64}$ & $\sfrac{-1}{2}$ & \\
\texttt{can} & grow & 1 & 0 & \\
\texttt{can} & move & $\sfrac{1}{4}$ & $\sfrac{-1}{2}$ & \\
\texttt{can} & swim & $\sfrac{1}{16}$ & $\sfrac{-5}{8}$ & \\
\texttt{can} & fly & $\sfrac{1}{16}$ & $\sfrac{-1}{8}$ & \\
\texttt{can} & sing & $\sfrac{1}{64}$ & $\sfrac{-17}{32}$ & \\
\texttt{has} & bark & $\sfrac{1}{16}$ & 0 & \\
\texttt{has} & petals & $\sfrac{1}{16}$ & $\sfrac{-1}{2}$ & \\
\texttt{has} & wings & $\sfrac{1}{16}$ & $\sfrac{-1}{8}$ & \\
\texttt{has} & feathers & $\sfrac{1}{16}$ & $\sfrac{-1}{8}$ & \\
\texttt{has} & scales & $\sfrac{1}{16}$ & $\sfrac{-5}{8}$ & \\
\texttt{has} & gills & $\sfrac{1}{16}$ & $\sfrac{-5}{8}$ & \\
\texttt{has} & roots & $\sfrac{1}{4}$ & 0 & \\
\texttt{has} & skin & $\sfrac{1}{4}$ & $\sfrac{-1}{2}$ & \\
\bottomrule
\end{tabular}
\end{table}

\paragraph{Handling leaves.} 
As mentioned above, \textit{almost} all plants have leaves, except pine, which has needles instead of leaves. How can we accommodate this exception? 
It is of course possible to follow the same reasoning as when we added colors: append one extra dimension to the representation space, and associate the relation/attribute pair ``has leaves'' to all entities/concepts who have leaves: flower, oak, rose, and daisy. 
However, we describe a different strategy which shows how we can model exceptions to a rule. 
We start by encoding the (erroneous) rule that all plants have leaves, which can be easily done by adding to Table~\ref{tab:quillian_padic_relations} a new entry ``has leaves'' with the same weights  as the entry for ``has roots'': 
($w_k = \sfrac{1}{4}$, $w_k'=0$, and $v_{kj}=0$). 
Now we need to create exceptions for this rule: since a pine has no leaves, we can no longer say that any of its ancestors has leaves; therefore we have to create exceptions for the entities/concepts pine, tree, and plant. 
This can be done by adding a third dimension to the representation space, call it $x_i''$, which is 0 for entities/concepts which are neither pine, tree, or plant, and which are 1 for any of those entities/concepts. Next we choose a weight $w_k''$ which  multiplies this new feature, which is 0 for all attributes except ``leaves''---this has no effect on entities outside the exception list, but changes the decision rule for pine, tree, and plant to 
\begin{align}
    \hat{y}_k(i, j) = \left\{
    \begin{array}{ll}
         +1 & \text{if $|w_{k} x_i + w_{k}'' x_i'' + v_{kj}|_p \le 1$} \\
         -1 & \text{otherwise,}
    \end{array}
    \right. 
\end{align}
when $j,k$ correspond to ``has leaves''. 
Since by design all these entities $i$  (pine, tree, and plant) satisfy $w_k x_i + v_{kj} \in \mathbb{Z}_p$, any $w_k'' \notin \mathbb{Z}_p$ (\textit{e.g.},  $w_k''=\sfrac{1}{2}$) creates the desired exceptions.

\begin{figure}
    \centering
    \includegraphics[width=0.6\columnwidth]{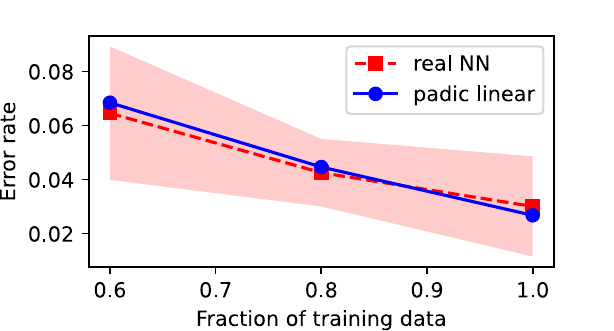}
    \caption{Experiment with a learned neural network reproducing \citet{rumelhart1990brain} and a linear $p$-adic network with embeddings set as in Table~\ref{tab:quillian_padic_entities} and whose weights $\{w_k\}$ and $\{v_{kj}\}$ are learned through Algorithm~\ref{alg:padic_beam_search}. Shown are test error rates when \{60\%, 80\%, 100\%\} of the training data is used.}
    \label{fig:quillian_padic_exp}
\end{figure}

\paragraph{Experiment.}  
We run a simple synthetic experiment using the semantic network above as follows. 
We generate all possible 1,680 propositions combining the 15 entities, 4 relations, and 28 outputs illustrated in Figure~\ref{fig:semantic_network}. 
We randomly split this full dataset into 80\% training propositions and 20\% test propositions. 
We further sample reduced versions of the training set with 60\% and 80\% examples. 
We implement a neural network (``real NN'') as \citet{rumelhart1990brain} with a 6-dimensional embeddings and a 15-dimensional hidden layer (690 $\mathbb{R}$ parameters) and a linear $p$-adic network (``padic linear'') with 3-dimensional embeddings (241 $\mathbb{Q}_p$ parameters) chosen as in Table~\ref{tab:quillian_padic_entities}, both illustrated in Figure~\ref{fig:semantic_network}. 
We train the neural network with gradient backpropagation using the Adam optimizer (learning rate 0.1), and the linear $p$-adic classifier with  Algorithm~\ref{alg:padic_beam_search}. 
We run 10 trials with different random initializations for the former and a single trial for the latter, since it is deterministic. 
Both models managed to overfit the full training set (in all trials) with zero error rate. 
The results in Figure~\ref{fig:quillian_padic_exp} show the results for the train/test split using different training set sizes. 
We observe that the two models generalize similarly to the test set.  

\section{Open Problems}\label{sec:open_problems} 

This paper only touches the surface of how one might perform machine learning in $p$-adic spaces by establishing basic theoretical results. 
To find out whether this framework might be practically useful, many challenges have yet to be surpassed. 
I summarize below some open problems and suggest possible paths to research them. 

\subsection{Linear multi-class $p$-adic classifiers}\label{sec:multi-class}

In \S\ref{sec:classification} we addressed only one-class and binary classification problems. 
In multi-class problems, we can have $K \ge 2$ classes. 
While any multi-class problem can be reduced to a combination of binary problems (\textit{e.g.} through one-against-all or one-against-one schemes), it is interesting to try to derive a native framework for $p$-adic multi-class classification. 

We could define a weight vector $w_k \in \mathbb{Q}_p^d$ and bias $b_k  \in \mathbb{Q}_p$ for each output class $k \in [K]$ and have a decision rule like 
$$\hat{y}(x) = \arg\min_{k \in [K]} |w_k^\top x + b_k|_p.$$ 
A potential drawback with this approach is that ties are very likely, due to the discrete nature of $p$-adic norms. 
Let us think about the unidimensional case where $d=1$ and assume $|w_k|_p=1$ for all $k$ to simplify. In this case, by defining ``class centroids'' $a_k := -b_k/w_k$, the problem becomes that of returning the nearest centroid for a given $x$. 
The ``Voronoi cells'' associated to each class are arranged hierachically as a dendrogram, and there might be ties since the space is ultrametric. 
Note that not all ties are permitted: there can only be a tie involving two classes if any other classes that are more similar to either of the two are also in the tie.  

\subsection{Gradient-based optimization and $p$-adic root finding}

While we provide efficient classification and regression algorithms for the unidimensional case, for $d>1$ the learning algorithms presented here (Algorithm~\ref{alg:padic_beam_search} for $p$-adic classification and the algorithm sketched in \S\ref{sec:linear_regression_multidimensional} for $p$-adic regression) should be seen merely as a proof of concept, since they are not practical---they take exponential runtime with respect to the number of features $d$. Finding better algorithms is an open problem, on which the relevance of $p$-adic predictors for any practical purpose strongly depends upon. 

One open question is whether the tools of $p$-adic calculus can lead to better learning algorithms---indeed, many of the tools from real analysis are also available to the $p$-adic numbers, including formal power series, continuity, and differentiation \citep{koblitz1984padic,robert2000course}. However, doing this requires  overcoming several roadblocks---while Newton's algorithm works in the $p$-adics to find roots of $p$-adic functions based on formal derivatives, $p$-adic calculus is fundamentally different for real calculus: 
\textit{e.g.}, a function with zero derivative everywhere might not be constant in $\mathbb{Q}_p$. 
Furthermore, the classical idea in machine learning of optimizing a loss function to fit a model to the training data does not work in a straightforward way in the $p$-adics---since $\mathbb{Q}_p$ is not an ordered field, ``optimizing'' a $p$-adic valued loss is meaningless. In our regression formulation in \S\ref{sec:regression}, we bypassed this by optimizing a $p$-adic \textit{norm}, which is $\mathbb{Q}$-valued. 
It is likely that many interesting $p$-adic learning problems can be mathematically formulated as $p$-adic min-norm problems, a topic which deserves further investigation. 

We should note, however, that a cornerstone of $p$-adic analysis is Hensel's lemma, an analogous to Newton's method for root finding which has strong convergence properties. This suggests that a possible path for learning $p$-adic predictors is to ingeniously design a function $F$ leading to a $p$-adic equation $F(\theta; \mathcal{D}) = 0$, where $\theta \in \mathbb{Q}_p^d$ are the model parameters and $\mathcal{D}$ is the training data, in a way that a solution of this equation corresponds to a ``optimal'' model configuration in some sense. In machine learning models over the reals one would choose $F(\theta; \mathcal{D}) = \nabla L(\theta; \mathcal{D})$ for a differentiable loss function $L$, but in the $p$-adic case it would be convenient to work directly with $F$, bypassing the loss function. 
Finding a root of $F$ through $p$-adic Newton's algorithm by generalizing Hensel lifting lemma \citep[\S 4]{gouvea2020padic} to $\mathbb{Q}_p^d$ would likely lead to very efficient algorithms. 

\subsection{Multi-layer $p$-adic networks}

In this paper, we have resorted to linear predictors (with the exception of the higher-order classifiers covered in \S\ref{sec:nonlinear_classifiers}). However, in machine learning over the reals, multi-layer networks can be much more expressive \citep{hornik1989multilayer}. We naturally expect the same to happen with ``$p$-adic neural networks''. 
For example, while we have shown in Proposition~\ref{prop:properties_padic_classifiers} that linear $p$-adic classifiers cannot solve count thresholding problems, it is easy to construct a single hidden-layer $p$-adic neural network which solve such problems: 
we can simply form a first layer of $d-k+1$ $p$-adic classifiers solving \textit{exact} counting problems for $i = k, k+1, ..., d$---using the construction presented in Appendix~\ref{sec:proof_classifier_properties}---and then, noting that at most one of these $d-k+1$ classifiers returns $+1$ (the others must return $-1$), append a top layer with a single exact-one classifier with weights $w_1 = ... = w_{d-k+1} = p^{-n}$ and bias $b = (d-k-1)p^{-n}$, with $n = \lceil \log_p (d-k+1) \rceil$; the resulting multi-layer system returns $+1$ if at least $k$ inputs are ``true'' and $-1$ otherwise. 

To make progress in this research question, we need to seek suitable non-linearities and learning criteria. 

\subsection{Using all primes: adelic predictors}\label{sec:adelic_predictors}

Our whole paper assumes some prime $p$ is fixed from which a non-Archimedean norm $|\cdot|_p$ is constructed. However, which $p$ should be chosen? Can we build predictors which use multiple primes or even \textit{all} primes simultaneously? 

The question above has a similar flavor as the so-called Hasse's \textbf{local-global principle} \citep[\S 4.8]{gouvea2020padic}, a cornerstone of number theory.  
The idea behind this principle is to try to answer a ``global'' complex question in $\mathbb{Q}$ (\textit{e.g.} finding a rational solution of a Diophantine equation) by working simultaneously at all ``local'' completions, \textit{i.e.}, $\mathbb{Q}_p$ for each prime $p$ as well as $\mathbb{R}$. 
In fact, it is a consequence of Ostrowski's theorem \citep[Theorem 3.1.4]{gouvea2020padic} that any completion of $\mathbb{Q}$ has one of these forms: 
every non-trivial absolute value on $\mathbb{Q}$ is equivalent (in the sense of inducing the same topology) 
to one of the non-Archimedean $p$-adic absolute values $|\cdot|_p$ (Example~\ref{ex:padic_abs}) or to the usual Archimedean absolute value $|\cdot|_\infty$ (Example~\ref{ex:usual_abs}). 
Mathematicians often use $p=\infty$ (the ``prime at infinity'') to index the set of  real numbers, denoting $\mathbb{Q}_\infty = \mathbb{R}$. 
We can thus express all completions of $\mathbb{Q}$ as the set $\left\{ \mathbb{Q}_p  \,:\, p \le \infty\right\}$. 
An example of a local-global result is the  \textbf{product formula} \citep[Proposition 3.1.5]{gouvea2020padic}, which states that, for any $x \in \mathbb{Q} \setminus \{0\}$, 
\begin{align}\label{eq:product_formula}
\prod_{p \le \infty} |x|_p = 1.
\end{align}
(This formula is very easy to prove by using the fundamental theorem of arithmetic.)

I next sketch a path for \textbf{adelic classification} taking inspiration from this principle. 
Let us think about an ``ensemble'' of $p$-adic classifiers for all primes $p$ including also a ``real classifier'', denoted as $p=\infty$, whose decision rule---given an input $x \in \mathbb{Q}^d$---is given by
\begin{align}\label{eq:adelic_decision_rule}
\hat{y}(x) := \left\{
\begin{array}{ll}
+1, & \text{if $\prod_{p \le \infty} |w_{(p)}^\top x + b_{(p)}|_p 
\le 1$}\\
-1, & \text{otherwise.}
\end{array}
\right.
\end{align}
We can think of this decision rule as letting all $p$-adic classifiers (for each $p$) vote through their corresponding absolute value, using their specific parameters $w_{(p)} \in \mathbb{Q}_p^d$ and $b_{(p)} \in \mathbb{Q}_p$, and then collecting all the votes to produce the final decision. 
From the product formula \eqref{eq:product_formula}, we have that, if $w_{(p)} = w \in \mathbb{Q}^d$ and $b_{(p)} = b \in \mathbb{Q}$ (\textit{i.e.}, if the parameters are the same for all $p$) and $w^\top x + b \ne 0$, we have 
$\prod_{p \le \infty} |w_{(p)}^\top x + b_{(p)}|_p 
= \prod_{p \le \infty} |w^\top x + b|_p = 1$, \textit{i.e.}, all the predictions lie at the decision boundary. 
On the other hand, if we assume that only finitely many $p$ in a set $\bar{\mathcal{P}}$ contribute to the voting and all the others $p \notin \bar{\mathcal{P}}$ have $w_{(p)} = 0$ and $b_{(p)} = 1$, we can think of algorithms which progressively add new primes $p$ to improve model-data fit until some stop criterion is met. 

To parametrize the classifier \eqref{eq:adelic_decision_rule} we need to define $w = (w_{(p)})_{p \le \infty}$ and  $b = (b_{(p)})_{p \le \infty}$---these objects are called \textbf{adeles} \citep{goldfeld2011automorphic}. Formally, the ring of adeles $\mathbb{A}_{\mathbb{Q}}$ is defined (with elementwise addition and multiplication) as
\begin{align}\label{eq:adeles}
\mathbb{A}_{\mathbb{Q}} := \prod'_{p \le \infty} \mathbb{Q}_p = \left\{ z \in \prod_{p \le \infty} \mathbb{Q}_p \,:\, \text{$|z_{(p)}|_p \le 1$ for all but finitely many $p$} \right\},
\end{align}
where $\prod'$ denotes the \textit{restricted} product (rather than the Cartesian product), which requires all but finitely many entries of $z = (z_{(p)})_{p \le \infty} \in \mathbb{A}_{\mathbb{Q}}$ to be $p$-adic integers. 
Note that, for any $z \in \mathbb{A}_\mathbb{Q}$, the restricted product ensures that $[z] := \prod_{p \le \infty} |z|_p$ converges, which enables its use in \eqref{eq:adelic_decision_rule}. 
Note also that any rational number $q$ can be ``seen'' as an adele $z = (q, q, ...)$ with all entries constant, \textit{i.e.}, we have an injection $\mathbb{Q} \xhookrightarrow{} \mathbb{A}_{\mathbb{Q}}$. 
From the product formula \eqref{eq:product_formula}, we have that adeles corresponding to these rational numbers have $[z] = 1$. 
Also, any $a \in \mathbb{Q}_p$ can also be seen as an adele $z$ with $z_{(p)} = a$ and $z_{(p')} = 1$ for $p' \ne p$, for which we have $[z] = |z|_p$ (this works also for $p=\infty$, \textit{i.e.}, real numbers). 
Since adeles form a ring, we have that for each $x \in \mathbb{Q}^d$---or even $x \in \mathbb{A}_{\mathbb{Q}}^d$---$w \in \mathbb{A}_\mathbb{Q}^d$ and $b \in \mathbb{A}_\mathbb{Q}$ ensure that $z = w^\top x + b \in \mathbb{A}_\mathbb{Q}$ and we can write \eqref{eq:adelic_decision_rule} as 
$\hat{y} = +1$ if $[w^\top x + b] \le 1$ and $\hat{y} = -1$ otherwise. 
Note that the $p$-adic classifiers defined in \S\ref{sec:classification} are a particular case of this construction resulting from setting $w_{(p)} = 0$ and $b_{(p)} = 1$ for all but a particular $p$. 
We can also define \textbf{adelic regression} by considering residuals $[w^\top x + b - y]$, which are also a generalization of the $p$-adic regression framework presented in \S\ref{sec:regression}. 

Developing the concepts above might be an interesting path for future work. 

\section{Conclusions}

We presented an exploratory study of ``$p$-adic machine learning'', which replaces the field $\mathbb{R}$ by $\mathbb{Q}_p$. 
We established the main building blocks for $p$-adic classification and regression, including prediction rules, learning formulations, and learning algorithms. 
We derived foundational theoretical properties, some of which somewhat  surprising, a consequence of the ultrametricity of $\mathbb{Q}_p$: \textit{e.g.}, linear regressors are exemplar-based (they pass through $d+1$ training points); unidimensional linear classifiers are enclosing $p$-adic balls, and nonlinear classifiers with polynomial discriminant functions are unions of these balls. 
We showed that the topology of $\mathbb{Q}_p$ is appealing for capturing hierarchical relations between objects in the representation space, such as those arising in semantic networks, and we provided a proof of concept for a network designed by \citet{rumelhart1990brain} for which there is a compact linear network in $\mathbb{Q}_p$ with perfect accuracy, whereas this is not the case in $\mathbb{R}$. 

Non-Euclidean geometric representations, such as hyperbolic embeddings, have been studied to capture hierarchical relationships \citep{nickel2017poincare}.
Our framework is radically different from such approaches, which still use the
field of real numbers as a backbone. 
We believe our work is only a first step towards $p$-adic machine learning, and we identify many potential directions for future work. 
Overcoming these challenges is an exciting direction for future work.

\bibliographystyle{apalike}
\bibliography{references}

\newpage
\appendix


\section{In Ultrametric Spaces Every Triangle Is Isosceles}\label{sec:proof_isosceles}

Invoke the strong triangle inequality \eqref{eq:sti} and assume  $d(x,y) \ne d(y,z)$. We can assume without loss of generality that $d(x,y) > d(y,z)$. From \eqref{eq:sti} we have $d(x,z) \le \max\{d(x,y), d(y,z)\} = d(x,y)$, but applying \eqref{eq:sti} again we have $d(x,y) \le \max\{d(x,z), d(y,z)\}$, and since we assumed $d(x,y) > d(y,z)$ we must have $d(x,y) \le d(x,z)$. Therefore, combining the the two inequalities we must have $d(x,z) = d(x,y) = \max\{d(x,y), d(y,z)\}$. 

\section{Additional Properties of $p$-adic Balls}\label{sec:padic_balls_properties}

In the next examples, for arbitrary sets $A$ and $B$ (subsets of an additive group), we define the sets $a + B := \{a+b \,:\, b \in B\}$,  $cB := \{cb \,:\, b \in B\}$, and the (Minkowski) set sum $A+B := \{a+b \,:\, a
\in A, b \in B\}$. 
Note that last two operations are both distributive with respect to the union, \textit{i.e.}, 
$A+(B \cup C) = (A+B) \cup (A+C)$ and $c(A \cup B) = cA \cup cB$. 

\begin{example}
Any ball in $\mathbb{Q}_p$ with nonzero radius can be written as $\bar{B}_{p^n}(a) = a + p^{-n} \mathbb{Z}_p$. 
In particular, the decomposition \eqref{eq:Zp_partition} can be equivalently written self-referentially as 
\begin{align}
    \mathbb{Z}_p = \bigcup_{i=0}^{p-1} (i + p\mathbb{Z}_p).
\end{align}
This shows that $\mathbb{Z}_p$ has self-similar structure.
\end{example}

\begin{proposition}
The $p$-adic balls are closed under Minkowski sums and we have
\begin{align}
    \bar{B}_r(a) + \bar{B}_s(b) = \bar{B}_{\max\{r,s\}}(a+b).
\end{align}
The $p$-adic balls with Minkowski sums form a commutative monoid with $\{0\} = \bar{B}_0(0)$ as the identity.  
Furthermore, $p$-adic balls are also closed under scalar multiplication and we have, for any $c \in \mathbb{Q}_p$,
\begin{align}
    c \bar{B}_r(a) = \bar{B}_{r|c|_p}(ca).
\end{align}
\end{proposition}

\begin{proof}
    Note first that $a + \bar{B}_s(b) = \bar{B}_0(a) + \bar{B}_s(b) = \{a + y \,:\, |y-b|_p \le s\} = \{u \,:\, |u-a-b|_p \le s\} = \bar{B}_s(a+b)$. 
    Now, suppose (without loss of generality) that $r \ge s$. Since balls with the same center must be nested, we have $y \in \bar{B}_s(b) \Longrightarrow y \in \bar{B}_r(b)$, therefore, for any $a \in \mathbb{Q}_p$,  $|y-b|_p = |y-b+a-a|_p \le r$, that is $a+b \in \bar{B}_r(a+y)$. Since any point in a ball is a center of the ball, we must have $\bar{B}_r(a+y)=\bar{B}_r(a+b)$. 
    Therefore, we have 
    $\bar{B}_r(a) + \bar{B}_s(b) = \bigcup_{y \in \bar{B}_s(b)} ( \bar{B}_r(a) + y) = \bigcup_{y \in \bar{B}_s(b)} \bar{B}_r(a+y) = \bigcup_{y \in \bar{B}_s(b)} \bar{B}_r(a+b) = \bar{B}_r(a+b)$. 

    Regarding the second part, note that if $c=0$, the statement becomes $\{0\} = \bar{B}_0(0)$, which is true. Now assume $c\ne 0$. We have $c \bar{B}_r(a) = \{cx \,:\, |x-a|_p\le r\} = \{u \,:\, |\frac{u}{c}-a|_p\le r\} = \{u \,:\, |u-ca|_p\le r|c|_p\} = \bar{B}_{r|c|_p}(ca)$.
\end{proof}

We then have the following corollary, which generalizes \eqref{eq:Zp_partition}. 

\begin{corollary}\label{cor:ball_decomposition}
Any ball in $\mathbb{Q}_p$ satisfies:
\begin{align}
\bar{B}_{p^n}(a) &= \bigcup_{i=0}^{p-1} \bar{B}_{p^{n-1}}(a + p^{-n} i), 
\end{align}
where the union is disjoint, as well as $\bar{B}_{p^{n-1}}(a) = p \bar{B}_{p^n}(ap^{-1})$.
\end{corollary}

\begin{proof}
    Using the decomposition of the ball of $p$-adic integers, the proposition above, and the distributive properties of Minkowski sums and scalar multiplication with unions, we have $\bar{B}_{p^n}(a) = a + p^{-n} \mathbb{Z}_p = a + p^{-n} \bigcup_{i=1}^{p-1} (i + p\mathbb{Z}_p) = a + p^{-n} \bigcup_{i=1}^{p-1} \bar{B}_{p^{-1}}(i) = a + \bigcup_{i=1}^{p-1} p^{-n}  \bar{B}_{p^{-1}}(i) = \bigcup_{i=1}^{p-1}   \bar{B}_{p^{n-1}}(a + p^{-n} i)$. It is straightforward to see that this union is disjoint. The second statement is easily proved from the proposition above.
\end{proof}

\section{A $p$-adic Proof of Kraft's Inequality}\label{sec:kraft}

The connection expressed in Example~\ref{ex:padic_strings} between $p$-adic balls and strings can be used to obtain a simple proof of  \textbf{Kraft's inequality} \citep{kraft1949device}, an important result in information theory which establishes a necessary and sufficient condition for a code to be a prefix code (hence, uniquely decodable). 
\begin{proposition}[Kraft's inequality]\label{prop:kraft}
    Let each source symbol from the alphabet $S = \{s_1, \ldots, s_n\}$ 
be encoded into a prefix code over an alphabet of size $m$ with codeword lengths
$\ell_{1},\ell_{2},\ldots ,\ell_{n}$.
Then
$\sum_{i=1}^{n}m^{-\ell _{i}} \le 1$. 
Conversely, for a given set of natural numbers 
$\ell_{1},\ell_{2},\ldots ,\ell_{n}$ 
satisfying the above inequality, there exists a prefix code over an alphabet of size 
$m$ with those codeword lengths. 
\end{proposition}

We provide a simple proof for the case where $m = p$ is prime based on properties of $p$-adic balls. 


We start with the following lemma. 

\begin{lemma}\label{lemma:ball_decomposition}
    Let $\{\bar{B}_{r_i}(a_i)\}_{i=1}^n$ be a finite set of disjoint balls in $\mathbb{Q}_p$, \textit{i.e.}, satisfying $\bar{B}_{r_i}(a_i) \cap \bar{B}_{r_j}(a_j) = \varnothing$ for $i\ne j$. Assume $r_i = p^{n_i}$ for $n_i \in \mathbb{Z}$. 
    Let $\bar{B}_{r}(a)$ be another ball enclosing all the balls $\bar{B}_{r_i}(a_i)$, where $r = p^n$. Then, we must have $\sum_{i=1}^n r_i \le r$, with equality iff $\bar{B}_r(a) = \bigcup_{i=1}^n \bar{B}_{r_i}(a_i)$.
\end{lemma}

\begin{proof}
    From Corollary~\ref{cor:ball_decomposition}, we have that $\bar{B}_{p^n}(a)$ can be decomposed as a disjoint union of $p$ smaller balls, each with radius $r_i = p^{n-1}$. Then we have $\sum_{i=0}^{p-1} p^{n-1} = p^n$. We can proceed recursively by decomposing some of the smaller balls further, which will keep the equality. Since all balls in $\mathbb{Q}_p$ must be either nested or have empty intersection, these are the only possible ways to obtain a decomposition of a ball into smaller balls. The set $\{\bar{B}_{r_i}(a_i)\}_{i=1}^n$ is necessarily a subset of one of these decompositions where some balls may be missing, hence the sum $\sum_{i=1}^n r_i$ is upper bounded by $r = p^n$.
\end{proof}

We now provide a simple proof of Proposition~\ref{prop:kraft} for the case where $m := p$ prime, using the facts we already know about $p$-adic balls. 
We start with the 
the $\Longrightarrow$ direction.  
We identify each string (code) with a ball contained in $\mathbb{Z}_p$;  
a string with length $\ell_i$ corresponds to a ball with radius $p^{-\ell_i}$ and some center $a_i \in \mathbb{Z}_p$ such that the first $\ell_i$ digits of $a_i$ correspond to the characters of the string. 
Since this is a prefix code, none of the balls corresponding to a codeword can be included in another ball corresponding to a different codeword, \textit{i.e.}, the set of balls $\{\bar{B}_{p^{-\ell_i}}(a_i)\}$ have pairwise empty intersection. Hence, it satisfies the conditions of Lemma~\ref{lemma:ball_decomposition} and therefore, since $\mathbb{Z}_p = \bar{B}_1(0)$ encloses all these balls, we must have $\sum_{i=1}^n p^{-\ell_i} \le 1$. 
To prove the $\Longleftarrow$ direction, suppose that $\ell_1$, ..., $\ell_n$ satisfy $
\sum_{i=1}^n m^{-\ell_i} \le 1$ and assume without loss of generality that $\ell_1 \le ... \le \ell_n$. 
We can create a prefix code by first picking a string (code) associated with the ball $\bar{B}_{p^{-\ell_1}}(0)$, then another string (code) associated with a ball with radius $p^{-\ell_2}$ which is not included in the first ball, etc.

\section{Proofs}

\subsection{Proof of Proposition~\ref{prop:binary_classifier_conditions}}\label{sec:proof_binary_classifier_conditions}

To show 1, note that $w^\star \xx{i} + b^\star = {\xx{i}}/(\xx{j} - \xx{i}) - {\xx{i}}/(\xx{j} - \xx{i}) = 0$ and $w^\star \xx{j} + b^\star = {\xx{j}}/(\xx{j} - \xx{i}) - {\xx{i}}/(\xx{j} - \xx{i}) = 1$. 
For every $x_+ \in \mathcal{D}_+$, we have 
$|w^\star x_+ + b^\star|_p = {|x_+ - \xx{i}|_p}/{|\xx{j} - \xx{i}|_p} \le 1$ from the definition of $\xx{j}$. 
For every $x_- \in \mathcal{D}_-$, we have 
$|w^\star x_- + b^\star|_p = {|x_- - \xx{i}|_p}/{|\xx{j} - \xx{i}|_p} > 1$, since $\mathcal{D}$ is assumed linearly separable, and therefore, due to ultrametricity, the distance between any two positive examples must be strictly smaller that the distance between a positive and a negative example. 

To see 2, note that $a=-\frac{b^\star}{w^\star} = \xx{i}$ and $r=\frac{1}{|w^\star|_p} = |\xx{j} - \xx{i}|_p$. 

To show 3, note the classifier defined by $\bar{B}_{r'}(a)$ corresponds to $a = -\frac{b}{w}$ and $r' = \frac{1}{|w|_p}$ for some $w$ and $b$; the second equation is satisfied with $b = -wa = -w\xx{i}$. We have, for any $x_- \in \mathcal{D}_-$,  $|wx_- + b|_p = |w(x_- + \xx{i})|_p = |w|_p |x_- + \xx{i}|_p = \frac{1}{r'}|x_- + \xx{i}|_p \ge pr' / r' = p > 1$. 
For any $x_+ \in \mathcal{D}_+$,  $|wx_+ + b|_p = |w(x_+ + \xx{i})|_p = |w|_p |x_+ + \xx{i}|_p = \frac{1}{r'}|x_+ + \xx{i}|_p < pr' / r' = p$, due to linear separability and ultrametricity. Since $|wx_+ + b|_p < p$, we must have $|wx_+ + b|_p \le 1$. 

Point 4 follows automatically from points 1 and 3.


\subsection{Proof of Proposition~\ref{prop:second_order}}\label{sec:proof_second_order}


    Let us first prove 1. We start by showing the ``$\Rightarrow$'' direction, \textit{i.e.}, that any $x$ satisfying $|f(x)|_p \le 1$ is of the form 
    \begin{align}\label{eq:second_order_ball_disjunction}
        x \in \bar{B}_{p^{-k+k_{12}}}(a_1) \cup \bar{B}_{p^{-k+k_{12}}}(a_2),
    \end{align}   
    We have  that $|f(x)|_p \le 1$ is equivalent to $|x-a_1|_p|x-a_2|_p \le r = p^{-k}$. Letting $p^{-k_1} = |x-a_1|_p$ and $p^{-k_2} = |x-a_2|_p$, this holds iff 
    \begin{align}\label{eq:second_order_ball_disjunction_proof_01}
    k_1 + k_2 \ge k. 
    \end{align}
    Moreover, from the strong triangle inequality, we have that $p^{-k_{12}} = |a_1 - a_2|_p \le \max\{|x - a_1|_p, |x - a_2|_p\} = \max \{p^{-k_1}, p^{-k_2}\}$, hence $k_{12} \ge \min\{k_1, k_2\}$, with equality if $k_1 \ne k_2$. 
    We now note that, in the scenario 1 where $k>2k_{12}$, we must have $k_1 \ne k_2$: indeed, if $k_1 = k_2$ we would have $k_{12} \ge k_1 = k_2$ and therefore $k > 2k_{12} \ge k_1+k_2$, which contradicts  \eqref{eq:second_order_ball_disjunction_proof_01}. As a consequence, we have $k_{12} = \min\{k_1, k_2\}$. Suppose $k_1 < k_2$ (\textit{i.e.} $|x - a_1|_p > |x - a_2|_p$).  Then, $k_{12} = k_1$ and, combining with  \eqref{eq:second_order_ball_disjunction_proof_01}, we get $k_2 \ge k - k_{12}$, which implies $|x-a_1|_p = p^{-k_{12}}$ and 
    \begin{align}\label{eq:second_order_ball_disjunction_proof_02}
    |x-a_2|_p \le p^{-k+k_{12}}.
    \end{align}
    But since $k>2k_{12}$ we have $-k+k_{12} < -k_{12}$, and therefore \eqref{eq:second_order_ball_disjunction_proof_02} implies $|x-a_2|_p < p^{-k_{12}}$; from the strong triangle inequality and since $|a_1-a_2|_p = p^{-k_{12}}$, we have automatically $|x-a_1|_p = p^{-k_{12}}$, which shows that \eqref{eq:second_order_ball_disjunction_proof_02} alone is always a feasible solution. 
    If we had supposed above that $k_1 > k_2$, by symmetry we would have obtained $|x-a_1|_p \le p^{-k+k_{12}}$, which together with \eqref{eq:second_order_ball_disjunction_proof_02} shows that the disjunction \eqref{eq:second_order_ball_disjunction} holds in scenario 1. 
    Conversely, let $x$ be of the form \eqref{eq:second_order_ball_disjunction}. if $x \in \bar{B}_{p^{-k+k_{12}}}(a_1)$, we have as shown above that $|x-a_2|_p = p^{-k_{12}}$, therefore $f(x) = |c|_p|x-a_1|_p|x-a_2|_p \le p^k p^{-k+k_{12}} p^{-k_{12}} = 1$, which proves the ``$\Leftarrow$'' direction. 
 
    Let us now prove 2, again starting with the ``$\Rightarrow$'' direction. Let $x$ satisfy $|f(x)|_p \le 1$ with $k \le 2k_{12}$. Letting again $p^{-k_1} = |x-a_1|_p$ and $p^{-k_2} = |x-a_2|_p$, we  have again \eqref{eq:second_order_ball_disjunction_proof_01} as well as the inequality $k_{12} \ge \min\{k_1, k_2\}$, with equality if $k_1 \ne k_2$. If $k_1 < k_2$, we can use a similar reasoning as above to obtain \eqref{eq:second_order_ball_disjunction_proof_02} conjoined with $|x-a_1|_p = p^{-k_{12}}$; but now we have $-k + k_{12} \ge -k_{12}$, and therefore the strong triangle inequality implies $|x-a_2|_p \le p^{-k_{12}} = |x-a_1|_p$. 
    We obtain the similar result $|x-a_1|_p \le p^{-k_{12}} = |x-a_2|_p$ if we assume $k_1 > k_2$. The third possibility to consider is $k_1 = k_2$. In this case the strong triangle inequality implies $k_{12} \ge k_1 = k_2$, so we must have $|x-a_1|_p = |x-a_2|_p \ge p^{-k_{12}}$, and we must also have from \eqref{eq:second_order_ball_disjunction_proof_01} that $k_1 = k_2 \ge \lceil k/2 \rceil$, which implies $|x-a_1|_p = |x-a_2|_p \le p^{-k/2}$. Since $\lceil k/2 \rceil \le k_{12}$, putting everything together we obtain $x \in \bar{B}_{p^{-\lceil k/2 \rceil}}(a_1) = \bar{B}_{p^{-\lceil k/2 \rceil}}(a_2)$ as desired. 
    Finally, to prove the ``$\Leftarrow$'' direction, note that any $x$ as above satisfies $|x-a_1|_p \le p^{-\lceil k/2 \rceil}$ and $|x-a_2|_p \le p^{-\lceil k/2} \rceil$, therefore $f(x) = |c|_p |x-a_1|_p |x-a_2|_p \le p^k p^{-\lceil k/2 \rceil} p^{-\lceil k/2 \rceil} \le 1$.

\subsection{Proof of Proposition~\ref{prop:properties_padic_classifiers}}\label{sec:proof_classifier_properties}

\paragraph{Boolean problems with $d=2$.}  
We start by showing that, for $d=2$ and any prime $p$, classifiers in $\mathcal{H}_{\mathbb{Q}_p^d}$ can compute any Boolean function. 
We assume for convenience that $(x_1, x_2) \in \{0, 1\}^2$ (we could as well assume $(x_1, x_2) \in \{-1, +1\}^2$ and we would obtain the same proof with a linear transformation of the weights). 
There are 16 cases to consider---$2^4$ possible output assignments for the 4 input configurations. 
The always-zero and always-one classifiers can be easily solved by picking $w=0$ and choosing a suitable $b$; the cases where the output depends only on $x_1$ or only on $x_2$ are also easy since we can choose respectively $w_2=0$ or $w_1=0$ to ignore the irrelevant input and revert to a unidimensional problem. 
Negating one of the inputs can also be handled with a transformation $x_i' = 1-x_i$ and recalculating $w$ and $b$ accordingly. 
The only cases left to analyze are:
\begin{enumerate}
    \item The XOR function. Here the target is $y(x) = +1$ if $x_1 \ne x_2$ and $-1$ otherwise. We can set $w_1 = w_2 = \frac{1}{p}$ and $b=-\frac{1}{p}$. We get $|w^\top x + b|_p = 0 \le 1$ if $x_1 \ne x_2$, $|w^\top x + b|_p = \big|\frac{1}{p}\big|_p = p > 1$ if $x_1=x_2$.  
    \item The negation of the XOR function. Similar to the previous case; we can set $w_1 = -w_2 = \frac{1}{p}$ and $b=0$. 
    \item The AND function. Here the target is $y(x) = +1$ if $x_1 = x_2 = 1$ and $-1$ otherwise. For $p=2$, we can set $w_1 = w_2 = \frac{1}{4}$ and $b=-\frac{1}{2}$. For $p>2$, we can set $w_1 = w_2 = \frac{1}{p}$ and $b=-\frac{2}{p}$. We will see below that, like XOR, this is a particular case of a counting problem for which we provide a general solution.  
    \item The negation of the AND function. This cannot be solved directly. If it were, we would need to have $|b|_p \le 1$, $|w_1 + b|_p \le 1$, $|w_2 + b|_p \le 1$, and $|w_1 + w_2 + b|_p > 1$. By the strong triangle inequality, we have $|w_1 + w_2 + b|_p \le \max\{|w_1 + b|_p, |w_2|_p\}$, so to satisfy the inequalities above we would need to have $|w_2|_p > 1$. 
    Therefore $|w_2|_p \ne |b|_p$, and hence we must have $|w_2 + b|_p = \max\{|w_2|_p, |b|_p\} = |w_2|_p > 1$, which contradicts the assumption. We will see below that this is a particular case of a count thresholding problem. 
    However, since the class $\mathcal{H}_{\mathbb{Q}_p^d}$ is defined to contain both classifiers $\hat{y}(.; w, b)$ and their negations $-\hat{y}(.; w, b)$, we can revert to the previous problem.  
\end{enumerate}

\paragraph{Congruence, parity checks, and counting problems.} 
We next show that, for $d\ge 2$, $\mathcal{H}_{\mathbb{Q}_p^d}$ can solve congruence problems modulo $p^n$ (of which parity check problems are a special case, when $p=2$ and $n=1$) as well as and counting problems. 
We assume inputs are binary, $(x_1, ..., x_d) \in \{0, 1\}^d$. 
Congruence  modulo $p^n$ problems correspond to the following target function: 
\begin{align}\label{eq:congruence_target}
    y(x) = \left\{
    \begin{array}{ll}
    +1 & \text{if $\sum_{i=1}^d x_i \equiv a \,\,\mathrm{mod}\,\, p^n$}\\
    -1 & \text{otherwise},
    \end{array}
    \right.
\end{align}
where $n \in \mathbb{N}$ and $a \in \{0, ..., p^{n}-1\}$. 
These problems are solved by a $p$-adic linear classifier with $w_i = p^{-n}$, for each $i \in [d]$, and $b = -ap^{-n}$. 
When $c$ inputs are active, we obtain $|w^\top x + b|_p = |(c - a)p^{-n}|_p = p^n |c-a|_p$, which is $\le 1$ iff $|c-a|_p \le p^{-n}$, that is, $c \equiv a \,\,\mathrm{mod}\,\, p^n$. 

Counting problems (of which XOR and AND are particular cases) correspond to the following target function (again with domain $\{0, 1\}^d$):
\begin{align}\label{eq:counting_target}
    y(x) = \left\{
    \begin{array}{ll}
    +1 & \text{if $\sum_{i=1}^d x_i = c$}\\
    -1 & \text{otherwise},
    \end{array}
    \right.
\end{align}
where $c \in \{0, ..., d\}$. 
These problems are solved by a $p$-adic linear classifier with $w_i = p^{-n}$, for each $i \in [d]$, and $b = -cp^{-n}$, with $n$ set as 
$n = \lceil \log_p (1 + \max\{c, d-c\}) \rceil$. To see this, observe that counting problem \eqref{eq:counting_target} is equivalent to congruence problem \eqref{eq:congruence_target} for $a=c$ and sufficiently large $n$. 
More specifically, $n$ should be large enough so that $|(c'-c)p^{-n}|_p > 1$ for any $c' \in \{0, ..., c-1, c+1, ..., d\}$. This inequality is equivalent to  $|c'-c|_p > p^{-n}$, which is ensured if $\max\{c, d-c\} < p^n$, or $n \ge \lceil \log_p (1 + \max\{c, d-c\}) \rceil$. 

\paragraph{Count thresholding problems.} 

This corresponds to the following target function (again with domain $\{0, 1\}^d$):
\begin{align}\label{eq:counting_thres_target}
    y(x) = \left\{
    \begin{array}{ll}
    +1 & \text{if $\sum_{i=1}^d x_i \le c$}\\
    -1 & \text{otherwise},
    \end{array}
    \right.
\end{align}
where $c \in \{0, ..., d\}$. 
The AND problem seen above is a particular case, when $d=2$ and $c=1$, and we already saw that it cannot be solved by $p$-adic linear classifiers for any $p$. 
In fact, the same happens for general $d \ge 2$ and $c \in \{1, ..., d-1\}$. 
We prove this by contradiction. 
Suppose there is $w$ and $b$ which solves \eqref{eq:counting_thres_target} for some $c \in \{1, ..., d-1\}$. 
Then we must have (in particular) $|b|_p \le 1$,  $|\sum_{j=1}^c w_j + b|_p \le 1$, $|w_{c+1} + b|_p \le 1$, 
and $|\sum_{j=1}^c w_j + w_{c+1} + b|_p > 1$. 
By the strong triangle inequality, we have $|\sum_{j=1}^c w_j + w_{c+1} + b|_p \le \max\{|\sum_{j=1}^c w_j + b|_p, |w_{c+1}|_p\}$, so to satisfy the inequalities above we would need to have $|w_{c+1}|_p > 1$. Therefore, $|w_{c+1}|_p \ne |b|_p$, and hence we must have  $|w_{c+1} + b|_p = \max\{|w_{c+1}|_p, |b|_p\} = |w_{c+1}|_p > 1$, which contradicts the assumption.

\subsection{Proof of Proposition~\ref{prop:classifier_conditions}}\label{sec:proof_classifier_conditions}

For 1, we need to show that for any $\xx{i}$ such that $|w^\top \xx{i} + b|_p \le 1$ we have (i) $|w^\top x + b|_p \le 1 \Rightarrow |w^\top x - w^\top \xx{i}|_p \le 1$ and (ii) $|w^\top x + b|_p > 1 \Rightarrow |w^\top x - w^\top \xx{i}|_p > 1$. 
Note that $|w^\top x - w^\top \xx{i}|_p = |w^\top x + b - w^\top \xx{i} - b|_p \le \max\{|w^\top x + b|_p, |w^\top \xx{i} + b|_p\}$, with equality if $|w^\top x + b|_p \ne |w^\top \xx{i} + b|_p$. 
For (i), since both $|w^\top x + b|_p \le 1$ and $|w^\top \xx{i} + b|_p \le 1$, it follows that $|w^\top x - w^\top \xx{i}|_p \le 1$. For (ii), since $|w^\top x + b|_p > 1$, we must have $|w^\top x + b|_p \ne |w^\top \xx{i} + b|_p$, and therefore $|w^\top x - w^\top \xx{i}|_p = \max\{|w^\top x + b|_p, |w^\top \xx{i} + b|_p\} = |w^\top x + b|_p > 1$. 

Now we prove 2. Let $w$ and $b$ be such that the classifier is tight and satisfies $w^\top \xx{i} + b = 0$ for some $i$ (guaranteed from point 1), and choose $j$ such that $|w^\top \xx{j} + b|_p = 1$. Then we have an invertible $u \in \mathbb{Z}_p^\times$ such that $u(w^\top \xx{j} + b) = 1$. Setting $w^\star = uw$ and $b^\star = ub$ and noting that ${w^\star}^\top \xx{i} + b^\star = u(w^\top \xx{i} + b) = 0$ completes the proof.

\subsection{Proof of Proposition~\ref{prop:unidimensional_regression}}\label{sec:proof_prop_unidimensional_regression}

We will make use of the following lemma. 

\begin{lemma}\label{lemma:exemplar}
    Let $a_1, \ldots, a_n \in \mathbb{Q}_p$. Then, for any $u \in \mathbb{Q}_p$ there is a $j \in [n]$ such that $|a_j - a_i|_p \le |u - a_i|_p$ for all $i \in [n]$. Concretely, any $j \in \arg\min_k |u - a_k|_p$ satisfies this bound. 
\end{lemma}

\begin{proof}
Let $j \in \arg\min_k |u - a_k|_p$ as stated, \textit{i.e.}, we have $|u-a_j|_p \le |u-a_k|_p$ for all $k$. 
Then, for any $i$, we have 
$|a_j - a_i|_p = |a_j - u - a_i + u|_p \le \max\{|u - a_j|_p, |u - a_i|_p\} = |u - a_i|_p.$
\end{proof}

It is instructive to examine first what happens in the zero-dimensional case, where $\xx{i} = 0$ for all $i = 1, \ldots, n$. In this case, only $\yy{1}, ..., \yy{n}$ matter, and the problem is that of finding the ``centroid'' $b$ that minimizes $\max_i |b - \yy{i}|_p$. Lemma~\ref{lemma:exemplar} implies that $b$ is a solution to this problem iff it is a point in the ball  spanned by $\yy{1}, ..., \yy{n}$ (hence, due to ultrametricity, also a center of that ball). In particular, $b = \yy{j}$ (for any $j \in [n]$) is a solution, again due to Lemma~\ref{lemma:exemplar}. 

We now consider another special case where there is no bias parameter, \textit{i.e.}, where the problem is to find $w \in \mathbb{Q}_p$ which minimizes $L(w) := \max_i |w\xx{i} - \yy{i}|_p$. 
Assume that $\xx{i} \ne 0$ for all $i \in [n]$. 
Then we have $L(w) = \max_i |\xx{i}|_p |w - \yy{i}/\xx{i}|_p$. 
From Lemma~\ref{lemma:exemplar}, we have that, for any $w$, there is a $j$ (namely $j \in \arg\min_k |w - \yy{k}/\xx{k}|_p$) such that $|\xx{i}|_p |\yy{j}/\xx{j} - \yy{i}/\xx{i}|_p \le |\xx{i}|_p |w - \yy{i}/\xx{i}|_p$ for all $i$, hence we can assume without loss of generality that the optimal $w$ is of the form $w = \yy{j}/\xx{j}$ for some $j$. 
We need to find the indices $i$ and $j$ associated to the min-max problem
$$\min_j \max_i |\xx{i}|_p \left|\frac{\yy{j}}{\xx{j}} - \frac{\yy{i}}{\xx{i}}\right|_p,$$
which can be done by explicit enumeration in $\mathcal{O}(n^2)$ time. 

Finally, let us address the case where we have a weight and a bias, \textit{i.e.}, we want to solve the problem
\begin{align}\label{eq:unidimensional_regression}
\min_{w, b} \max_i |w\xx{i} + b  - \yy{i}|_p.
\end{align}
Letting $w^\star$ be a partial solution to this problem, we have from the zero-dimensional case that any $b$ of the form $b^\star = \yy{k} - w^\star \xx{k}$ (for arbitrary $k$) completes the solution. 
Hence, we can substitute $b = \yy{k} - w\xx{k}$ and solve for $w$, which leads to 
$\min_{w} \max_i |w(\xx{i}-\xx{k}) - \yy{i}+\yy{k}|_p$. Since for $i=k$ we have  $|w(\xx{i}-\xx{k}) - \yy{i}+\yy{k}|_p = |0|_p = 0$, this problem is equivalent to $\min_{w} \max_{i\ne k} |w(\xx{i}-\xx{k}) - \yy{i}+\yy{k}|_p$. 
This now reverts to the problem without bias, from which we obtain the algorithm stated in Proposition~\ref{prop:unidimensional_regression}.

\subsection{Proof of Proposition~\ref{prop:regression_exemplar}}\label{sec:proof_regression_exemplar}

Let $w$ be a regressor which passes through $r<d+1$ points. We show that there is another regressor $\hat{w}$ passing through $r+1$ points such that $\|X\hat{w} - y\|_p \le \|Xw - y\|_p$. The result will follow by induction. 

Assume $w^\top \xx{i} = \yy{i}$ for $i \in \{j_1, ..., j_r\}$ and $w^\top \xx{i} \ne \yy{i}$ for $i \in [n] \setminus  \{j_1, ..., j_r\}$. 
Let $j_{r+1}, ..., j_d$ index arbitrary distinct inputs. 
We will show that there is $k \in [n] \setminus \{j_1, ..., j_d\}$ and a classifier $\hat{w}$ such that:
\begin{itemize}
    \item $\hat{w}^\top \xx{i} = w^\top \xx{i} = \yy{i}, \quad i \in \{j_1, ..., j_r\}$;
    \item $\hat{w}^\top \xx{i} = w^\top \xx{i} \notin \yy{i}, \quad i \in \{j_{r+1}, ..., j_d\}$;
    \item $\hat{w}^\top \xx{k} = \yy{k}$;
    \item $|\hat{w}^\top \xx{i} - \yy{i}|_p \le \max_{j \in [n]} |w^\top \xx{j} - \yy{j}|_p, \quad i \notin \{j_{r+1}, ..., j_d, k\}$. 
\end{itemize}
These conditions imply that $\|X\hat{w} - y\|_p \le \|Xw - y\|_p$. 

Define $\bar{X} \in \mathbb{Q}_p^{(d+1)\times (d+1)}$ as the submatrix formed by the rows $\{j_1, ..., j_d, k\}$ of $X$---this matrix is invertible by assumption. 
Likewise, let $\hat{y} \in \mathbb{Q}^{d+1}$ be  defined as $\yyhat{i} = w^\top \xx{i} = \hat{w}^\top \xx{i}$ for $i \in \{j_1, ..., j_d\}$ and $\yyhat{d+1} = \yy{k}$. 
We have $\bar{X}\hat{w} = \hat{y}$, and therefore $\hat{w} = \bar{X}^{-1} \hat{y}$. We further have, for $i \in [n]$:
\begin{align}\label{eq:proof_regression_what}
    \hat{w}^\top \xx{i} - \yy{i} &=  w^\top \xx{i} - \yy{i} + (\bar{X}^{-1}\hat{y} - w)^\top \xx{i}\nonumber\\
    &=  w^\top \xx{i} - \yy{i} + (\bar{X}^{-1}(\hat{y} - \bar{X}w))^\top \xx{i} \nonumber\\
    &=  w^\top \xx{i} - \yy{i} - (w^\top \xx{k} - \yy{k}) [\bar{X}^{-1}]^\top_{:, d+1} \xx{i}. 
\end{align}
We now note that matrix $\bar{X}$ can be decomposed as 
\begin{align}
    \bar{X} = \left[
    \begin{array}{cc}
        \tilde{X}_{1:d} & 1_d \\
        {\xxtildet{k}} & 1
    \end{array}
    \right],
\end{align}
where we denote by $\xxtilde{i} \in \mathbb{Q}_p^d$ the inputs before the bias augmentation for $i \in [n]$, \textit{i.e.}, $\xx{i} = [{\xxtildet{i}}, 1]^\top$, and we use $\tilde{X}_{1:d}$ to denote the matrix whose rows are ${\xxtildet{j_1}}, ..., {\xxtildet{j_d}}$. 
From the block matrix inversion formula---which is applicable since $\tilde{X}_{1:d}$ is invertible and $\bar{X}$ is also invertible, the latter implying that the Schur complement $(1 - \xxtildet{k} \tilde{X}_{1:d}^{-1} 1_d)$ is invertible---we obtain
\begin{align}
    [\bar{X}^{-1}]^\top_{:, d+1} \xx{i} &= \left[
    \begin{array}{c}
         -\tilde{X}_{1:d}^{-1} 1_d (1 - \xxtildet{k} \tilde{X}_{1:d}^{-1} 1_d)^{-1}  \\
         (1 - \xxtildet{k} \tilde{X}_{1:d}^{-1} 1_d)^{-1} 
    \end{array}
    \right]^\top \left[
    \begin{array}{c}
         \xxtilde{i}  \\
         1 
    \end{array}
    \right] \nonumber\\
    &= \frac{1 - \xxtildet{i} \tilde{X}_{1:d}^{-1} 1_d}{1 - \xxtilde{k} \tilde{X}_{1:d}^{-1} 1_d}.
\end{align}
We now define the following choice rule for $k$:
\begin{align}
    k := \argmax_{i \in [n] \setminus \{j_1, ..., j_d, k\}} \big|1 - \xxtilde{i} \tilde{X}_{1:d}^{-1} 1_d\big|_p, 
\end{align}
which ensures that $\big|[\bar{X}^{-1}]^\top_{:, d+1} \xx{i}\big|_p \le 1$. 

Finally, we apply the strong triangle inequality to \eqref{eq:proof_regression_what}:
\begin{align}
    |\hat{w}^\top \xx{i} - \yy{i}|_p &= \big|w^\top \xx{i} - \yy{i} - (w^\top \xx{k} - \yy{k}) [\bar{X}^{-1}]^\top_{:, d+1} \xx{i}\big|_p \nonumber\\
    &\le \max\left\{ 
    |w^\top \xx{i} - \yy{i}|_p, \,\,
    |w^\top \xx{k} - \yy{k}|_p \underbrace{\big| [\bar{X}^{-1}]^\top_{:, d+1} \xx{i}\big|_p}_{\le 1}  
    \right\} \nonumber\\
    &\le \max\left\{ 
    |w^\top \xx{i} - \yy{i}|_p, \,\,
    |w^\top \xx{k} - \yy{k}|_p
    \right\} \nonumber\\
    &\le \max_{j \in [n]} 
    |w^\top \xx{j} - \yy{j}|_p,
\end{align}
for all $i \in [n] \setminus \{j_1, ..., j_d, k\}$, which concludes the proof.

\end{document}